%% file: dependency-dependence.tex
\documentclass[11pt,table]{article}

\usepackage[pagenums]{emnlp2021}

\usepackage{times,latexsym}
\usepackage{url}

\usepackage[T1]{fontenc}

\usepackage[utf8]{inputenc}

\usepackage{microtype}

\usepackage{authblk} 
\makeatletter
\renewcommand\maketitle{\AB@maketitle} 
\renewcommand\AB@affilsepx{\quad\protect\Affilfont} 
\makeatother

\usepackage{xcolor}
\usepackage{booktabs}
\usepackage{graphicx}
\usepackage{mathtools}
\usepackage{soul}
\usepackage{amsfonts}
\usepackage{amsthm}
\usepackage{multirow}
\usepackage{afterpage}
\usepackage[linewidth=1pt]{mdframed}

\include{tikz-forest-parameters}

\newcommand{\pmi}{\mathrm{pmi}}
\newcommand{\CPMI}{\mathrm{CPMI}}

\title{Linguistic Dependencies and Statistical Dependence}

\author[1,2]{\textbf{Jacob Louis Hoover}}
\author[3]{\textbf{Alessandro Sordoni}}
\author[4]{\textbf{Wenyu Du}}
\author[1,2,5]{\textbf{Timothy J. O'Donnell}}

\affil[1]{McGill University}
\affil[2]{Mila Québec AI Institute}
\affil[3]{Microsoft Research, Montréal\authorcr}
\affil[4]{The University of Hong Kong}
\affil[5]{Canada CIFAR AI Chair, Mila\authorcr}
\affil[ ]{%
  \texttt{jacob.hoover@mail.mcgill.ca},\quad%
  \texttt{alsordon@microsoft.com},\authorcr%
  \texttt{wenyudu@connect.hku.hk},\quad%
  \texttt{timothy.odonnell@mcgill.ca}%
  }

\newcommand\fm\textsl
\newcommand\ex\textit

\date{}

\begin{document}
\maketitle

\begin{abstract}
Are pairs of words that tend to occur together also likely to stand in a
linguistic dependency? This empirical question is motivated by a long history
of literature in cognitive science, psycholinguistics, and NLP\@. In this
work we contribute an extensive analysis of the relationship between
linguistic dependencies and statistical dependence between words. Improving on
previous work, we introduce the use of large pretrained language models to
compute contextualized estimates of the pointwise mutual information between
words (CPMI). For multiple models and languages, we extract dependency trees
which maximize CPMI, and compare to gold standard linguistic dependencies.
Overall, we find that CPMI dependencies achieve an unlabelled
undirected attachment score of at most $\approx 0.5$.  While far above chance,
and consistently above a non-contextualized PMI baseline, this score is
generally comparable to a simple baseline formed by connecting adjacent words.
We analyze which kinds of linguistic dependencies are best captured in CPMI
dependencies, and also find marked differences between the estimates of the
large pretrained language models, illustrating how their different training
schemes affect the type of dependencies they capture.
\end{abstract}

\section{Introduction}

\begin{figure}
  \centering
  \includegraphics[width=\linewidth]{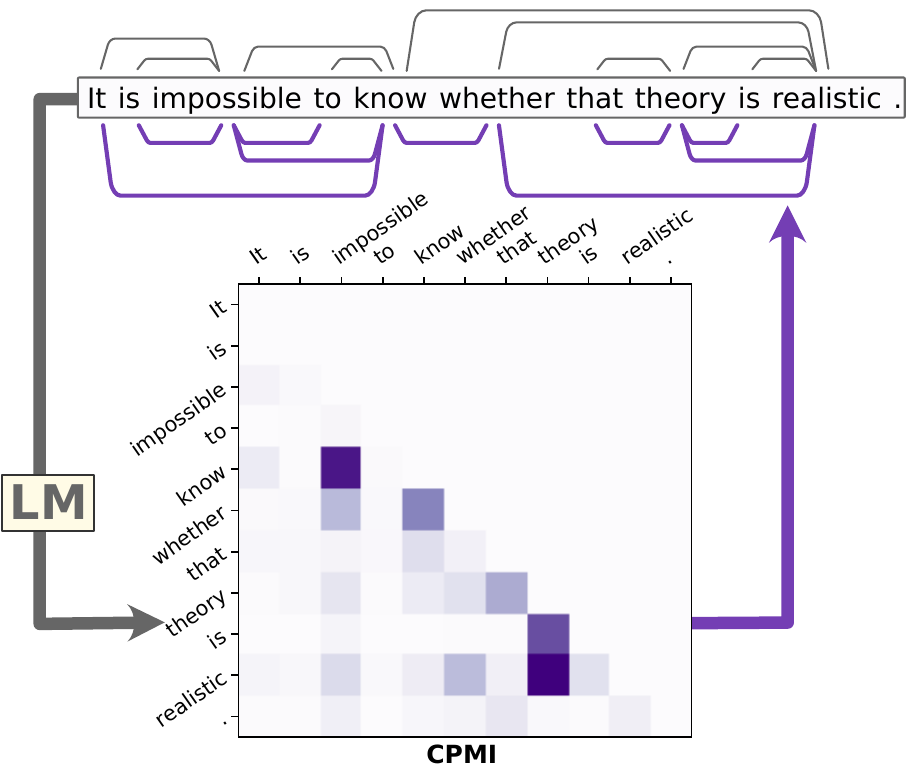}
  \caption{We use models pretrained on \fm{masked language modelling} objectives
  to extract trees which maximize contextualized pointwise mutual information
  (CPMI) between words, to examine how linguistic dependencies
  relate to statistical dependence.}%
  \label{fig:overview}
\end{figure}

A fundamental aspect of natural language structure is the set of \fm{dependency
relations} which hold between pairs of words in a sentence. Such dependencies
indicate how the sentence is to be interpreted and mediate other aspects of its
structure, such as agreement. Consider the sentence: \ex{Several ravens flew out
of their nests to confront the invading mongoose}. In this example,
there is a dependency between the verb \ex{flew} and its subject \ex{ravens},
capturing the role this subject plays in the flying event, and how it controls
number agreement. All modern linguistic theories recognize the centrality of
such word-word relationships, despite considerable differences in detail in how
they are treated
\citep[for a review of linguistic dependency grammar literature
see][]{marneffe.m:2019}.

In addition to linguistic dependencies between words, there are also clear and
robust statistical relationships. A noun like \ex{ravens} is likely to occur
with a verb like \ex{flew}. In short, the presence or absence of certain words
in certain positions in a sentence is informative about the presence or absence
of certain other words in other positions. This raises the question: Do words
that are strongly statistically dependent tend to be those related by linguistic
dependency (and vice versa)? In everyday language, a sentence like the example
above is probably more likely than \ex{Several pigs flew out of their nests to
confront the invading shrubbery}, despite this second example being
syntactically identical to the first. 

The long tradition of both supervised and unsupervised learning of grammars and
parsers in computational linguistics suggests a strong link between dependency
structure and statistical dependence. 
Works such as \citet{magerman.d:1990} and \citet{paiva-alves.e:1996} introduced
the use of pointwise mutual information (PMI) as a measure of the strength of
statistical dependence between words, for the purpose of inferring linguistic
structures from corpus statistics. The link between PMI and linguistic
dependency has been studied and affirmed in \citet{futrell.r:2019}. They show
that for words linked by linguistic dependencies, the estimated mutual
information between POS tags (and distributional clusters) is higher than that
between non-dependent word pairs, matched for linear distance.


In this work, we dig further into the question of the correspondence between
statistical and linguistic dependencies using modern pretrained language models
(LMs) to compute estimates of conditional PMI between words given context, which
we term \fm{contextualized pointwise mutual information} (CPMI). For each
sentence we extract a \fm{CPMI dependency tree}, the spanning tree with maximum
total CPMI, and compare these trees with gold standard linguistic dependency
trees.%
\footnote{We release our code at \url{https://github.com/mcqll/cpmi-dependencies}.}

We find that CPMI trees correspond better to gold standard trees than non context-dependent PMI
trees. However our analysis shows that CPMI dependencies and linguistic
dependencies correspond only roughly 50\% of the time, even when we introduce
a number of strong controls. Notably, we do not see better correspondence when
we examine CPMI trees inferred by models that are explicitly trained to recover
syntactic structure during training. Likewise, we see no increase in
correspondence when we calculate CPMI over part-of-speech (POS) tags, a control
designed to examine a less fine-grained statistical dependency than that between
actual word forms. In fact, CPMI arcs broadly correspond to linguistic
dependencies slightly less often than a simple baseline that just connects all
and only adjacent words. We see similar overall unlabeled undirected attachment
score (UUAS) when evaluated across a variety of pretrained models and different
languages. However, a close analysis shows noteworthy differences between the
different LMs, in particular revealing that BERT-based models are markedly more
sensitive to adjacent words than XLNet.  These differences yield insights about
how different LM pretraining regimes result in differences in how the models
allocate statistical dependencies between words in a sentence.

\section{Background}%
\label{sec:background-pmi}%
\fm{Pointwise mutual information} \citep[PMI;][]{fano.r:1961} is commonly used
as a measure of the strength of statistical dependence between two words.
Formally, PMI is a symmetric function of the probabilities of the outcomes $x,y$
of two random variables $X,Y$, which quantifies the amount of information about
one outcome that is gained by learning the other:
\[
  \pmi(x;y)\coloneqq
  \log\frac{p(x,y)}{p(x)p(y)}
  = \log\frac{p(x \mid y)}{p(x)}.
\]
In our case, the observations are two words in a sentence (drawn from discrete
random variables indexed by position in the sentence, ranging over the
vocabulary).
PMI has been used in computational linguistic studies as a measure of how words
inform each other's probabilities since \citet{church.k:1990}.\footnote{They
used the term \emph{word association}, which had a more subjective meaning in
the psycholinguistic literature, to refer specifically to PMI.}

Much earlier work on unsupervised dependency parsing
\citep[e.g.,][]{vandermude.a:1978, magerman.d:1990,carroll.g:1992,yuret.d:1998,
paskin.m:2002} used techniques involving maximizing estimates of total pointwise
mutual information between heads and dependents, or maximizing the conditional
probability of dependents given heads (these two objectives can be shown to be
equivalent under certain assumptions; see \S\ref{sec:hdpmi-hdcp}). While such
PMI-induced dependencies proved useful for certain tasks \citep[such as
identifying the correct modifier for a word among a selection of possible
choices;][]{paiva-alves.e:1996}, purely PMI-based dependency parsers did not
perform well at the general task of recovering linguistic structures
overall~\cite[see discussion in][]{klein.d:2004induction}.

The recent advent of pretrained contextualized LMs \citep[such as BERT,
XLNet;][]{devlin.j:2019, yang.z:2019} provides an opportunity to revisit the
relationship between PMI-induced dependencies and linguistic dependencies. These
networks are pretrained on very large amounts of natural language text using
masked language modelling objectives to be accurate estimators of conditional
probabilities of words given context, and thus are natural tools for
investigating the statistical relationships between words.

\section{Contextualized PMI dependencies}%
\label{sec:introducing-CPMI}

Linguistic dependencies are highly sensitive to context. For example, consider
the following two sentences: \ex{I see that the crows retreated}, and \ex{The
mongoose pursued by crows retreated}. In the first there is a dependency between
\ex{retreated} and \ex{crows}, and in the second there is not. However, PMI
between two words in a sentence is strictly independent of the other words in
that sentence.

\begin{figure}
  \centering
  \includegraphics[width=0.95\linewidth]{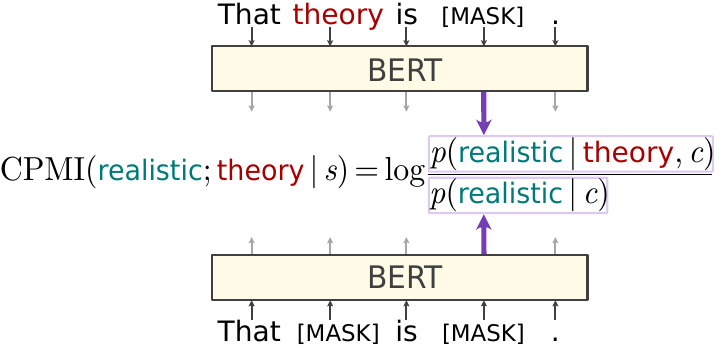}
  \caption{Diagram illustrating using BERT to compute the probability of
  \ex{realistic} with and without masking \ex{theory}, to obtain a CPMI score
  between those two words in the sentence $s=\ex{That theory is realistic}$.}%
  \label{fig:CPMI-bert}
\end{figure}

Here we define \fm{contextualized pointwise mutual information} (CPMI) as the
conditional PMI given context, which we estimate using pretrained contextualized
LMs.  A contextualized LM $M$ provides an estimate for the probability of words
given context, which we use to define CPMI$_M$ between two words $w_i$
and $w_j$ in a sentence $W$ as
\begin{align*}
  \CPMI_{M}(w_i;w_j) 
  &=\log\frac{p_M(w_i\mid W_{- i})}
    {p_M(w_i\mid W_{- i,j})}
\end{align*}
where the $W_{- i}$ is the sentence with word $w_i$ masked, and $W_{- i,j}$ is
the sentence with words $w_i,w_j$ masked. To demonstrate the computation of this
quantity, Figure~\ref{fig:CPMI-bert} illustrates how BERT is used to obtain a
CPMI score between the words \ex{theory} and \ex{realistic} in the sentence
\ex{That theory is realistic}.

\subsection{Dependency tree induction}%
\label{sec:extract-dependency-trees}

Given a sentence, we compute a matrix consisting of the CPMI between each pair
of words. We then symmetrize this matrix by summing across the diagonal, so that
we have a single score for each pair of words (omitting this step led to
extremely similar results).\footnote{Note that while theoretically CPMI should
  be symmetric, nothing in the pretraining of the LMs we use enforces this
identity (see~\S\ref{sec:symmetrizing-matrices} for details).} We then extract
tree structures which maximize total CPMI\@. Since natural language dependencies
are overwhelmingly projective \citep[see][]{kuhlmann.m:2010} we extract maximum
projective spanning trees using the dynamic programming algorithm from
\citet[][]{eisner.j:1996,eisner.j:1997}.\footnote{Eisner's algorithm recovers
  the optimal projective \textit{directed} dependency structure from a weighted
  ordered graph, but with a symmetric weight matrix, the output dependency trees
may be interpreted as undirected.}
Results for dependency trees alternatively extracted without the projectivity
constraint, using Prim's maximum spanning tree (MST) algorithm
\citep{prim.r:1957}, are similar, and results using both algorithms are provided
in \S\ref{sec:additional-results} for comparison. For further details on the
extraction of CPMI dependencies, see \S\ref{sec:details-calculating-CPMI}.

\begin{figure*}[h!]
  \centering
  \includegraphics[width=\linewidth]{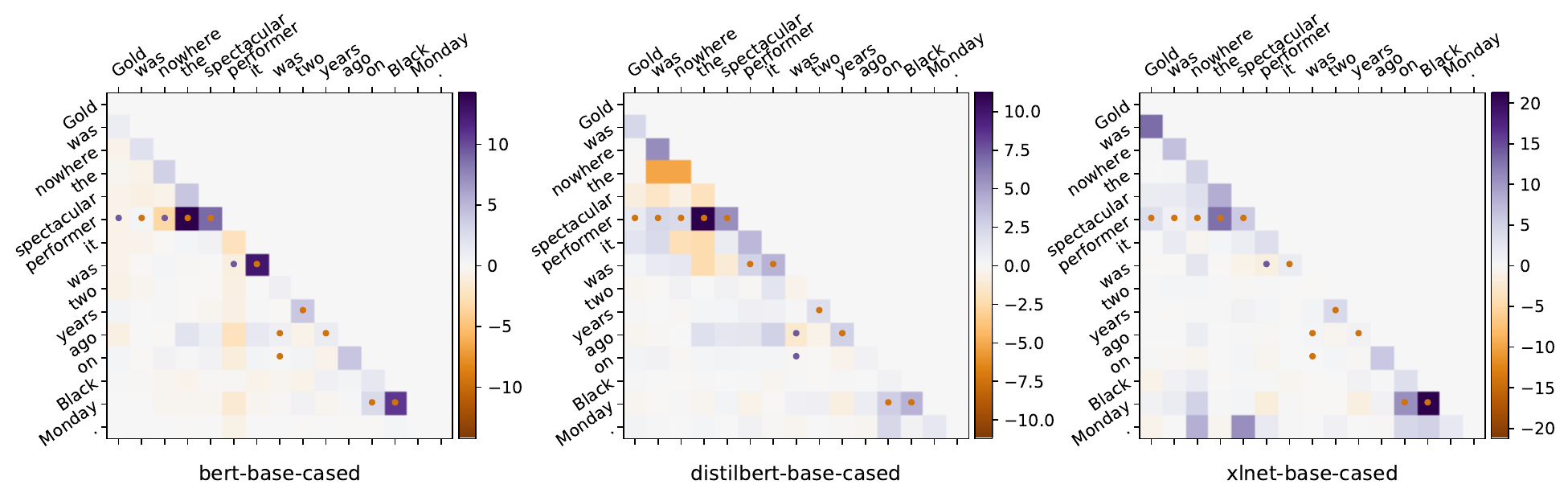}
  \begin{footnotesize}
    \input{figures/tikz/ce/1352sum.projective_BERT-large-wgold.tikz}\\
    \input{figures/tikz/ce/1352sum.projective_DistilBERT.tikz}\\
    \input{figures/tikz/ce/1352sum.projective_XLNet-base.tikz}\\
  \end{footnotesize}
  \caption{\textbf{Top:} CPMI matrices for an example sentence, from BERT,
DistilBERT, XLNet. Gold dependencies are marked with a dot. \textbf{Bottom:}
Resulting projective MST parses for the three models. Gold dependency parse
above in black, CPMI dependencies below, blue where they agree, and red when
they do not. The unlabeled undirected attachment score (UUAS) is given at
right. Further examples provided in appendix, Figure~\ref{fig:further-examples}%
.}%
\label{fig:example442}
\end{figure*}

\section{Evaluating CPMI dependencies}%
\label{sec:experiment-CPMI}

In this section, we analyze the degree to which CPMI-inferred dependencies from
pretrained LMs resemble linguistic dependencies.

\subsection{Method}
We use gold dependencies for sentences from the Wall Street Journal (WSJ), from
the Penn Treebank (PTB) corpus of English text hand-annotated for syntactic
constituency parses~\citep{marcus.m:1994}, converted into Stanford
Dependencies~\citep{demarneffe.m:2006stanforddep,demarneffe.m:2008}.%
\footnote{We use Stanford CoreNLP v3.9.2 to convert.} We evaluate all extracted
dependency trees on the full development split (WSJ section 22, consisting of
1700 sentences).  For comparison with other work in unsupervised grammar
induction, we also report results on the WSJ10 (all 389 sentences of length
$\le10$ from section 23, the test split, as used in e.g.~\citet{yang.s:2020})
in~\S\ref{sec:results-supplemental-ptb}.

To compare results across languages we use the Parallel Universal Dependencies
treebanks subset of Universal Dependencies~\citep[][v2.7]{nivre.j:2020ud2}.
These consist of 1000 sentences translated into 20 languages.

\paragraph{Pretrained contextualized LMs}
We compute CPMI scores using a number of transformer-based pretrained LMs for
English \citep[BERT, XLNet, XLM, BART, DistilBERT;][]{devlin.j:2019,
yang.z:2019, lample.g:2019, lewis.m:2019bart, sanh.v:2019distilbert}. For other
languages (and English) we use pretrained multilingual BERT base;
see~\ref{sec:results-supplemental-pud} for details. All pretrained
contextualized LMs we use are provided by Hugging Face
transformers~\citep{wolf.t:2020transformers}.

\paragraph{Syntactically aware models}
We likewise compute CPMI estimates using models explicitly designed to have a
linguistically-oriented inductive bias, by taking syntax into  account in their
training objectives and architecture. Following \citet{du.w:2020}, we include
two pretrained versions of an ordered-neuron LSTM~\citep[][]{shen.y:2018}---a
language model designed to have a hierarchical structural bias. The first
(ONLSTM) is pretrained on raw text data, the second (ONLSTM-SYD) is pretrained
on the same data but with an additional auxiliary objective to reconstruct PTB
syntax trees. As a control, we also include a vanilla LSTM model. All three
models are trained on the PTB training split. Example parses extracted from
these models are given in the appendix (Figure~\ref{fig:LSTM-examples}). We
extract CPMI estimates from these models similarly to the above, but we
condition only on preceding material, since these LSTM-based models operate
left-to-right. See \S\ref{sec:LtoR-CPMI} for details.%
\footnote{Note that results of the (ON)LSTM models are not directly
  comparable to the transformer-based models, 
  as these models are trained on much less data.}

\paragraph{Noncontextualized PMI control}
We also compute a non-contextualized PMI estimate using a pretrained global word
embedding model \citep[Word2Vec;][]{mikolov.t:2013}, to capture word-to-word
statistical relationships present in global distributional information, not
sensitive to the context of particular sentences. This control is calculated as the
inner product of Word2Vec's target and context embeddings,
\(\pmi_\textrm{w2v}(w_i;w_j) \coloneqq \mathbf{w}_i^\top \mathbf{c}_j \), since
its training objective is optimized when this quantity equals the PMI plus a
global constant~\citep[as explained in][]{levy.o:2014,allen.c:2019}. Details are
given in \S\ref{sec:w2v}.

\paragraph{Baselines} A random baseline is obtained by extracting a parse for
each sentence from a random matrix (so each pair of words is equally likely to
be connected). We also include a `connect-adjacent' baseline---degenerate trees
formed by simply connecting the words in order---a simple, strong, and
linguistically plausible baseline for English.

In addition to these baselines, we will compare unlabelled undirected accuracy
score (UUAS) with that reported for the Dependency Model with Valence
\citep[DMV;][]{klein.d:2004induction}, a classic dependency parsing model.
Note, importantly, the DMV is not fully unsupervised, as it relies on gold POS
tags, but it is still a useful benchmark, with UUAS 54.4\% on the entire WSJ
corpus, and 63.7\% on WSJ10 \citep[as reported
in][Fig.~3]{klein.d:2004induction}.

\subsection{Results}

\begin{table}[t]
  \small
  \centering
  \rowcolors{5}{}{lightgray!20}
  \begin{tabular}{llcc}
      {}                           & all & len $=1$ & len $>1$\\
      {}                           & {} & {\footnotesize prec. $|$ rec.}& {\footnotesize prec. $|$ rec.}\\
  \toprule
  random                           & .22 & .49 $|$ .34 & .08 $|$ .10\\
  connect-adjacent                 & \textbf{.49} & .49 $|$ \textbf{1}\phantom{.0} & \phantom{.0}-- $|$ 0\phantom{.0} \\
  \midrule
  \midrule
  Word2Vec                         & .39 & \textbf{.61} $|$ .59 & .19 $|$ .19\\                                 
  \midrule
  \textbf{BERT}  \small{base}      & .46 & .57 $|$ .72 & .27 $|$ .21\\
  \textbf{BERT}  \small{large}     & .47 & .55 $|$ .81 & .24 $|$ .13\\
  \textbf{DistilBERT}              & .48 & .57 $|$ .72 & \textbf{.32} $|$ .24\\
  \textbf{Bart}  \small{large}     & .38 & .52 $|$ .64 & .16 $|$ .13\\
  \textbf{XLM}                     & .42 & .60 $|$ .64 & .23 $|$ .22\\
  \textbf{XLNet} \small{base}      & .45 & .59 $|$ .66 & .29 $|$ \textbf{.25}\\
  \textbf{XLNet} \small{large}     & .41 & .59 $|$ .61 & .23 $|$ .22\\
  \midrule
  \midrule
  vanilla LSTM                     & .44 & .54  $|$ .70 & .26 $|$ .19\\  
  \midrule
  \textbf{ONLSTM}                  & .44 & .55  $|$ .71 & .27 $|$ .19\\  
  \textbf{ONLSTM-SYD}              & .45 & .55  $|$ .71 & .27 $|$ .19\\
  \bottomrule
  \end{tabular}
  \caption{Total UUAS for max-CPMI trees (projective). Overall scores in the
    first column (over all arcs in the corpus, precision $=$ recall), followed
    by precision and recall for adjacent words in the second and third columns,
    and likewise for nonadjacent words in the final two columns. Compare with an
    overall UUAS of \textbf{.544} originally reported in
    \citet{klein.d:2004induction} for the DMV on the WSJ corpus.}%
  \label{tab:cpmi-uuas}
\end{table}

\begin{table}[t]
  \small
  \centering
  \rowcolors{1}{}{lightgray!20}
  \begin{tabular}{lrrr}
    \toprule
    language    &  rand. & connect-adj. & \textbf{BERT} base  \\
    \midrule
    Chinese    &         .23 & .45 &   .40 \\
    Czech      &         .25 & .48 &   .48 \\
    English    &         .22 & .42 &   .43 \\
    French     &         .23 & .45 &   .47 \\
    German     &         .22 & .42 &   .46 \\
    Korean     &         .28 & .58 &   .49 \\
    Polish     &         .27 & .54 &   .52 \\
    Russian    &         .26 & .51 &   .51 \\
    Spanish    &         .23 & .45 &   .48 \\
    Turkish    &         .27 & .55 &   .48 \\
    \bottomrule
  \end{tabular}
  \caption{Total UUAS for selected languages from the multilingual Parallel UD
  dataset, for CPMI dependencies extracted from from BERT (base multilingual
  cased). See full results in Table~\ref{tab:pud_uuas-complete}.}%
  \label{tab:pud_uuas}
\end{table}

Example CPMI dependencies and extracted projective trees are given in
Figure~\ref{fig:example442}, with gold dependencies for comparison.
Table~\ref{tab:cpmi-uuas} gives the UUAS results.%
\footnote{The overall UUAS constitutes both precision and recall, since the
  number of gold edges and CPMI edges are the same: for a sentence of length
  $n$, the denominator is $n-1$.} Overall UUAS is given in the first column. The
  remaining columns give the UUAS for the subset of edges of length 1 and
  longer, in terms of precision and recall respectively.%
\footnote{For the connect-adjacent baseline, note: for length 1, the recall
  score is perfect, because all gold arcs of length 1 are predicted correctly by
  this trivial baseline; for the length $>1$ subset, precision is undefined
  since there are no predicted edges of length $>1$, and recall is 0.}
Table~\ref{tab:pud_uuas} gives overall UUAS from multilingual BERT for a
selection of languages from the PUD treebanks (for full results see
Table~\ref{tab:pud_uuas-complete}, Figure~\ref{fig:pud_uuas-complete}).

The overall results show broadly that CPMI dependencies correspond to linguistic
dependencies better than the noncontextual PMI-dependencies estimated from
Word2Vec. However, across the models, and across languages, UUAS in general is
in the range 40--50\%. Degenerate trees formed by connecting words in linear
order (the connect-adjacent baseline) achieve similar UUAS. Additionally, for
the ONLSTM models, which have a hierarchical bias in their design, we see that
accuracy of the CPMI-induced dependencies is the essentially the same with or
without the auxiliary syntactic objective. Overall accuracy for both
syntactically aware models is the same as for the vanilla LSTM\@.  Further
analysis of these results is in \S\ref{sec:cpmi-analysis}.

\section{Delexicalized POS-CPMI dependencies}%
\label{sec:experiment-pos} 

In this second experiment we estimate CPMI-dependencies over part-of-speech
(POS) tags, rather than words. In the unsupervised dependency parsing literature
there is an ample history of approaches making use of gold POS tags \citep[see
e.g.,][]{bod.r:2006,cramer.b:2007, klein.d:2004induction}. Additionally, a
traditional objection to the idea of deducing dependency structures directly
from cooccurrence statistics, beyond data sparsity issues, is the possibility
that ``actual lexical items are too semantically charged to represent workable
units of syntactic structure''~\citep[as phrased
by][p.3]{klein.d:2004induction}. That is, perhaps words' patterns of
co-occurrence contain simply too much information about factors irrelevant to
dependency parsing, so as to drown out the information that would be useful for
recovering dependency structure. According to this line of thinking, we might
expect linguistic dependency structure to be better related to the statistical
dependencies between the \emph{categories} of words, rather than lexical items
themselves. Thus a version of CPMI calculated over POS tags would be predicted
to achieve higher accuracy than the CPMI calculated over lexical item
probabilities above.

\begin{figure}
  \centering
  \includegraphics[width=0.8\linewidth]{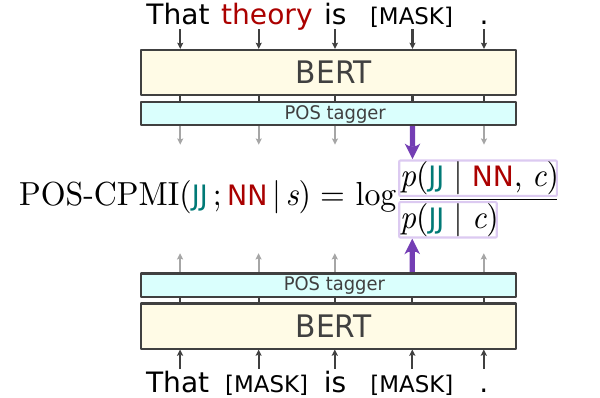}
  \caption{Diagram illustrating using BERT to compute the POS-CPMI score between
  the POS tags of the two words, \ex{theory} (a noun, \textsf{NN}) and
  \ex{realistic} (and adjective, \textsf{JJ}) in the sentence $s=\ex{That theory
  is realistic}$.}%
  \label{fig:CPMI-bert-pos}
\end{figure}

A straightforward but unfeasible way to investigate this idea would be to obtain
contextualized POS-embeddings by re-training all the LMs from scratch on large
delexicalized corpora only consisting of POS tags. Instead, for efficiency,
follow LM probing literature~\citep{hewitt.j:2019} and train a small POS probe
on top of a pretrained LM, which estimates the probability of the POS tag at a
given position in a sentence. After training this probe, we can extract a
POS-based CPMI score between words. We define this POS-CPMI analogously to
CPMI, but using conditional probabilities of POS tags, rather than word tokens:
\begin{align*}
&\mathrm{POS\text{-}CPMI}_M({\pi}_i;{\pi}_j) \\
&=
\log
\frac{p_{M_\mathrm{POS}}({\pi}_i \mid W_{-i})}
     {p_{M_\mathrm{POS}}({\pi}_i \mid W_{-i,j})}
\end{align*}
where $\pi_i,\pi_j$ are the gold POS tags of $w_i,w_j$ in sentence $W$, and
$M_\mathrm{POS}$ is the contextualized LM $M$ with a pretrained POS embedding
network on top. This is illustrated in Figure~\ref{fig:CPMI-bert-pos}. We then
extract POS-CPMI dependencies to compare to gold dependencies.

\subsection{Method}
We implement a POS probe as a linear transformation on top of the final hidden
layer of a fixed pretrained LM\@. We train two versions of this probe: one trained
simply to minimize cross entropy loss (simple POS probe), the other trained
using the information bottleneck
technique~\citep[following][]{tishby.n:2000,li.x:2019}, to maximize accuracy
while minimizing extra information included in the representation (IB POS
probe).  Using LMs BERT and XLNet (both base and large, each), we train each
type of probe, to recover PTB gold POS tags. All eight probes achieve between
92\% and 98\% training accuracy.

We extract parses from POS-CPMI matrices just for CPMI (described above in
\S\ref{sec:experiment-CPMI}). Below, we refer to the estimates extracted using
the simple POS probe as simple-POS-CPMI, and those extracted using the IB POS
probe as IB-POS-CPMI\@.

\subsection{Results}

\begin{table}
  \centering
  \rowcolors{3}{}{lightgray!20}
  \begin{tabular}{rllcc} 
  {}&         & all & len $= 1$ & len $> 1$  \\
  {}&         & {} & {\footnotesize prec. $|$ rec.}& {\footnotesize prec. $|$ rec.}\\
  \toprule 
  \cellcolor{white}&\textbf{BERT}  \small{base}   & .48 & .56 $|$ .79& .32 $|$ .19\\
  \cellcolor{white}&\textbf{BERT}  \small{large}  & .45 & .53 $|$ .75& .27 $|$ .16\\
  \cellcolor{white}&\textbf{XLNet} \small{base}   & .36 & .55 $|$ .56& .17 $|$ .17\\
  \multirow{-4}{*}{\rotatebox[origin=c]{90}{simple-POS}}
  \cellcolor{white}&\textbf{XLNet} \small{large}  & .32 & .56 $|$ .51& .14 $|$ .15\\
  \midrule
  \cellcolor{white}&\textbf{BERT}  \small{base}   & .41 & .58 $|$ .65 & .20 $|$ .18\\
  \cellcolor{white}&\textbf{BERT}  \small{large}  & .41 & .55 $|$ .69 & .18 $|$ .14\\
  \cellcolor{white}&\textbf{XLNet} \small{base}   & .40 & .55 $|$ .60 & .22 $|$ .20\\
  \multirow{-4}{*}{\rotatebox[origin=c]{90}{IB-POS}}
  \cellcolor{white}&\textbf{XLNet} \small{large}  & .36 & .56 $|$ .56 & .16 $|$ .16\\
  \bottomrule
  \end{tabular}
  \caption{Total UUAS for POS-CPMI using the simple POS probe and IB
  POS probe, from BERT and XLNet models. Overall results are in the first column,
  remaining columns break down results by arc length and recall and
  precision as in Table~\ref{tab:cpmi-uuas}.}%
  \label{tab:pos-cpmi-uuas}
  \end{table}

Using the POS-CPMI dependencies does not result in higher accuracy.
This provides evidence that the correlation between linguistic dependencies and
CPMI dependencies is not merely artificially low due to distracting lexical
information.

Table~\ref{tab:pos-cpmi-uuas} shows the UUAS of the simple-POS-CPMI and
IB-POS-CPMI trees. Compared to the lexicalized CPMI trees discussed in the
previous section, for BERT models, the simple-POS-CPMI dependencies have rather
comparable overall UUAS, while for XLNet it is markedly lower. For both models,
IB-POS-CPMI dependencies have lower UUAS\@. While these results are somewhat
mixed, it is clear that, in our experimental setting, POS-CPMI dependencies
correspond to gold dependencies no more than the CPMI dependencies do,
performing at best roughly as well as the connect-adjacent baseline.

\begin{figure*}[t]
    \centering
    \includegraphics[width=\linewidth]{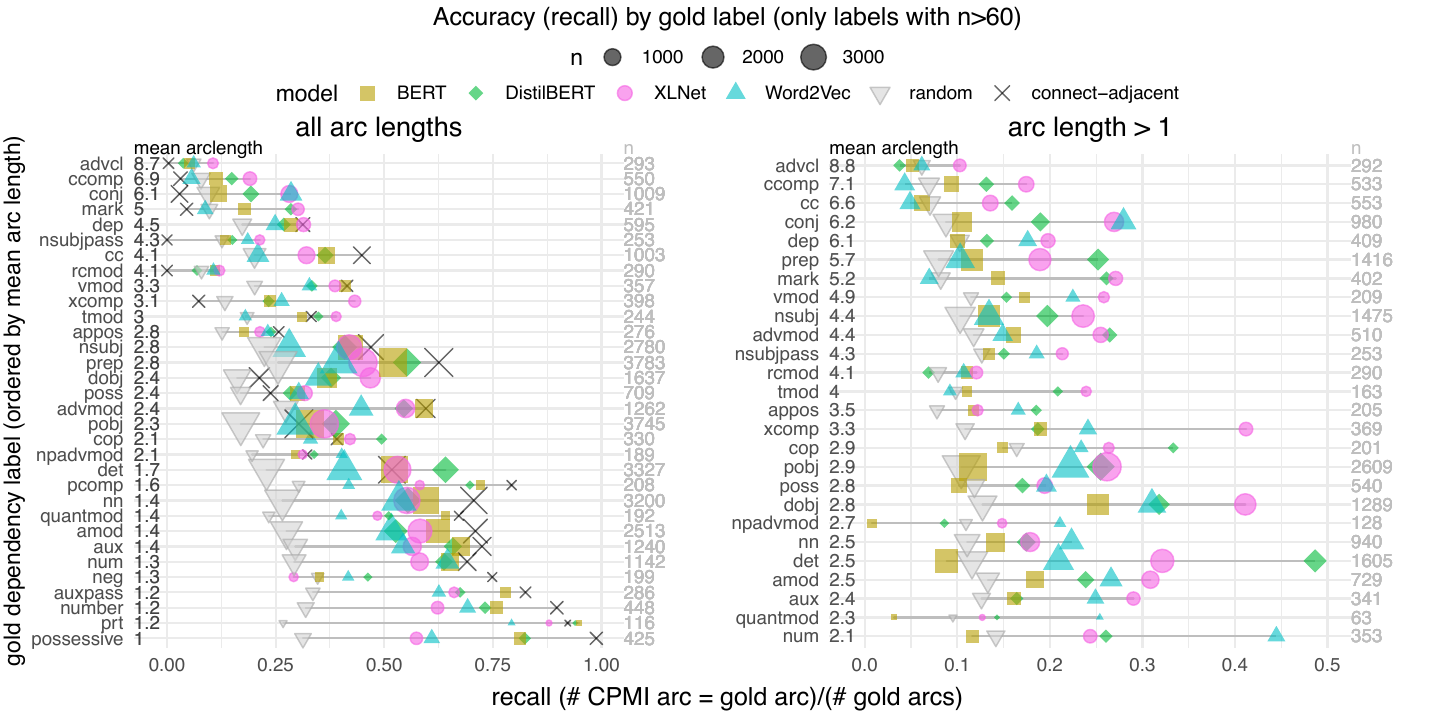}
    \caption{Plots of CPMI dependency recall accuracy versus gold edge relation
    (on the vertical axis, ordered by mean arc length). Only dependency
    relations of which there are more than 60 observations are included.
    \textbf{Left}: Including dependency arcs of all lengths. \textbf{Right}:
    Including only arcs between nonadjacent words. The connect-adjacent baseline
    predicts no such edges. Notice that the correlation with mean length
    disappears when excluding the length 1 arcs.}%
    \label{fig:relation}
\end{figure*}

\section{Analysis}%
\label{sec:cpmi-analysis}

In this section we outline main takeaways from a more detailed
examination of the results from
\S\S\ref{sec:experiment-CPMI}--\ref{sec:experiment-pos}, including additional
analysis in \S\ref{sec:cpmi-additional-analyses}.

\paragraph{UUAS is higher for length 1 arcs} 
Breaking down the results by dependency length, Figure~\ref{fig:lindist} (in
appendix) shows the recall accuracy of CPMI dependencies, grouped by length of
gold arc. Length 1 arcs have the highest accuracy, and longer dependencies have
lower accuracy. This trend holds for CPMI from all LMs. For BERT large, in
particular, arcs of length 1 have recall accuracy of 80\%, while longer arcs are
near random. For XLNet, this trend is less pronounced.

\paragraph{No relation label has high UUAS}
In Figure~\ref{fig:relation}, recall accuracy is plotted against gold dependency
arc label.\footnote{For descriptions of labels see the Stanford Dependencies
manual \citep[][]{demarneffe.m:2008sdmanual}} When examining all lengths of
dependency together (left) recall accuracy would seem to be correlated with mean
arc length. But, filtering out all the gold arcs of length 1 (49\% of arcs), we
see that there is not a strong overall effect of arclength on mean accuracy for
lengths > 1.

For most dependency labels, CPMI accuracy from each of the models is above the
random baseline, but at or below to the connect-adjacent baseline. Exceptions to
this trend include dependency labels \textsf{dobj} (direct object),
\textsf{xcomp} (which connects a verb or adjective to the root of its clausal
complement). For wordpairs in these relations, CPMI estimates (XLNet in
particular) achieve higher accuracy than the baselines. However, even in these
cases, CPMI dependencies do not perform at a level that could be considered
successful for an unsupervised parser. This is contrary to what would be
expected if CPMI-dependencies were in a strong correspondence with linguistic
dependencies, even if this only held for certain types of linguistic dependency.

When considering arcs of length > 1, there is no dependency arc label which has
UUAS above 0.5 from any of the models. More complete results including the other
models not shown in Figure~\ref{fig:relation} are given in
Table~\ref{tab:uuas-label-comparison} (in appendix).

\begin{figure}
    \centering \includegraphics[width=\linewidth]{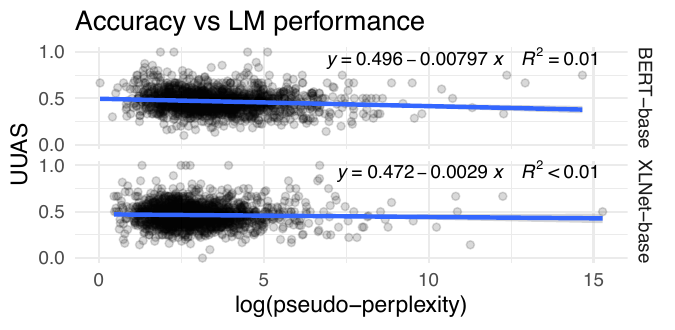}
    \caption{Per-sentence accuracy (UUAS) against log psuedo-perplexity.  Each
    dot represents one sentence. Fitting a linear regression, the coefficient of
    determination $R^2$ is very close to 0 for all models (here BERT and XLNet
    are shown; other models are in Figure~\ref{fig:uas-vs-ppl})}%
    \label{fig:uas-vs-ppl-small}
\end{figure}

\paragraph{UUAS is not correlated with LM performance}%
Figure~\ref{fig:uas-vs-ppl-small} shows per-sentence UUAS plotted against log
pseudo-perplexity (PPL) for BERT and XLNet models (results are similar for other
models; see~\S\ref{sec:acc-vs-ppl}, Figure~\ref{fig:uas-vs-ppl}).  These results
show that correspondence between CPMI-dependencies and linguistic dependencies
isn't higher on sentences on which the models are more confident.

We also examined the accuracy of CPMI dependencies during training of BERT (base
uncased) from scratch. Figure~\ref{fig:checkpoints} (in appendix) shows the
average perplexity of this model at checkpoints during training, along with
average UUAS of induced CPMI structures. UUAS reaches its highest value before
perplexity plateaus.

We should also stress that, throughout this paper, UUAS is not a measure of LM
quality.  Rather, it simply measures how well patterns of statistical dependence
captured by the LM align with linguistic dependencies.  Better alignment may not
be related to better language modelling.

\begin{figure}
    \centering
    \includegraphics[width=\linewidth]{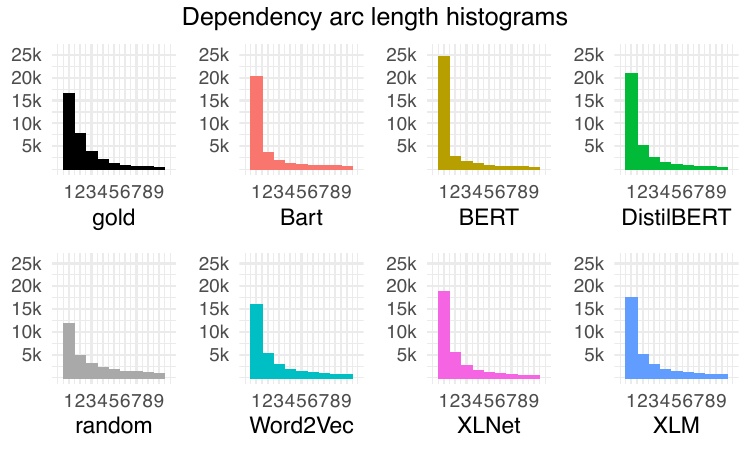}
    \caption{Histograms of arc length. Note, 49\% of the gold arcs are length 1,
      whereas all of the CPMI dependencies had a higher proportion. BERT (base),
      in particular has 72\%. For Word2Vec (which does not
      have access to word order), 47\% are length 1.
      For the connect-adjacent baseline (not shown) the histogram is
      trivial: all arcs are length 1.}%
    \label{fig:lindisthist}
\end{figure}

\paragraph{Dependencies differ between LMs}%
Dependency structures extracted from the different pretrained LMs show roughly
similar overall UUAS, though the models agree with each other on only 25--48\%
of edges.  They agree with the noncontextualized word embedding model Word2Vec
at just slightly lower rates (21--27\%), while agreeing with the linear baseline
at higher rates (34--57\%).  See \S\ref{sec:model-similarity} and for these
details.

In particular, CPMI dependencies from all the models connect adjacent words more
often than the gold dependencies do, but this effect is much more pronounced for
BERT models than for XLM, and XLNet models (Figure~\ref{fig:lindisthist}). A
possible reason for this difference lies in the way these models are trained.
XLNet is trained to predict words according to randomly sampled chain rule
decompositions, enforcing a bias to be able to predict words in any order,
including longer dependencies. XLNet's probability estimates for words may
therefore be sensitive to a larger set of words, rather than mostly the adjacent
ones. Whereas BERT, trained with a less constrained masked LM objective,
has probability estimates that are evidently more sensitive to adjacent words.

\section{Related work }%
\label{sec:related_work}

\paragraph{Probing pretrained embeddings} In the past few years, a substantial
amount of literature has emerged on probing pretrained language models \citep[in
the sense of e.g.][]{conneau.a:2018,manning.c:2020}, wherein a presumably weak
network (a \fm{probe}) is trained to extract linguistic information \citep[in
particular, dependency information, in e.g.][]{hewitt.j:2019,clark.k:2019} from
pretrained embeddings.
Extracting CPMI-dependencies differs from training a dependency probe in that it
is entirely unsupervised, and is motivated by a specific hypothesis---about
the relationship linguistic dependencies have with statistical dependence.

\paragraph{Nonparametric probing}
A number of other recent works have taken an unsupervised approach to
investigating syntactic structure encoded by pretrained LMs,
largely focusing on self-attention weights
\citep[e.g.][]{marecek.d:2018,marecek.d:2019,kim.t:2020pretrained,kim.t:2020chartbased,htut.p:2019}.
Very recently, \citet[][concurrent with this paper]{zhang.t:2021} examined
conditional dependencies implied by masked language modelling using a
nonparametric method similar to our CPMI, using BERT to estimate
Conditional PMI (and Conditional MI) between words. They extract maximum
spanning trees, and report UUAS on WSJ dependency data. Their results are
similar to those reported here: namely, scores are
much higher than a chance baseline, but close to a connect-adjacent
baseline. While their numerical results are similar, their interpretation
differs somewhat. Given our analysis, we find less reason for optimism
about the prospects of unsupervised dependency parsing directly from probability
estimates by pretrained LMs.

\paragraph{Perturbation impact} The experiments in the current paper extracting
CPMI can be seen as an application of the token perturbation
approach of~\citet{wu.z:2021}. They describe general nonparametric method to
examine the \fm{impact}, $f(w_i,w_j)$, of a word $w_j$ on another word $w_i$ in
the sentence, where $f$ is some difference function between the embedding of
$w_i$ (masked in the input) with and without the word $w_j$ also being masked. 
In their experiments, they use two examples of impact-measuring functions
\citep[see][\S2.2]{wu.z:2021}.  The first, the \fm{Dist} metric, is simply
Euclidean distance between embeddings. The second, the \fm{Prob} metric, is
defined as $f(w_i,w_j) = p(w_i\mid W_{-i}) - p(w_i\mid W_{-i,j})$, using the
masked LM's probability estimates (notation as defined in
\S\ref{sec:introducing-CPMI}).  The latter impact metric is quite similar to
CPMI, the difference being only that \fm{Prob} impact is the difference in
probabilities, while CPMI is the difference in log probabilities.

Table~\ref{tab:impact-cpmi-comparison} compares the reported UUAS of maximum
projective spanning trees from CPMI matrices, to those from \fm{Dist} impact
matrices on the English PUD data set. They do not report UUAS for the \fm{Prob}
metric or release code for it, but mention that it is significantly outperformed
by the \fm{Dist} method.  \citet[][p.1]{wu.z:2021} note that their ``best
performing method does not go much beyond the strong right-chain baseline''.
While it may be seen as an application of perturbed masking technique, CPMI is
motivated as a method to test a specific hypothesis about the relationship
between linguistic and statistical dependence.  Extracting matrices using
another impact metric (such as Euclidean distance between embeddings, \fm{Dist})
may indeed achieve higher attachment scores, as \citet{wu.z:2021} demonstrate,
but this does not bear on the hypothesis we focus on in this paper.

\section{Discussion}%
\label{sec:discussion-conclusion} 
In this paper we explored the connection between linguistic dependency and
statistical dependence.
We contribute a method to use modern pretrained language models to compute CPMI,
a context-dependent estimate of PMI, and infer maximum CPMI dependency trees
over sentences.

We find that these trees correlate with linguistic dependencies better than
trees extracted from a noncontextual PMI estimate trained on similar data.
However, we do not see evidence of a systematic correspondence between
dependency arc label and the accuracy of CPMI arcs, nor do we see evidence that
the correspondence increases when using models explicitly designed to encode
linguistically-motivated inductive biases, nor when estimated between
POS embeddings instead of word forms. Overall, CPMI-inferred dependencies
correspond to gold dependencies no more than a simple baseline connecting
adjacent words. This is our first main takeaway: statistical dependence (as
modelled by these pretrained LMs) is not a good predictor of linguistic
dependencies.
Second, our analysis shows that CPMI trees extracted from different LMs
differ to an extent that is perhaps surpising, given the similarity in spirit of their training
regimes. The difference in accuracy when broken down with respect to linear
distance between words offers information about the ways in which these models'
inductive and structural biases inform the way they perform the task of
prediction. BERT aligns better overall, but this is driven by its being more
like the linear baseline. For longer arcs, XLNet aligns a bit better with
linguistic structure. Compared to BERT, XLNet can be seen as imposing a
constraint on the language modelling objective by forcing the model to have
accurate predictions under different permutation masks.

Generalizing this observation, we ask whether linguistic dependencies would
correspond to the patterns of statistical dependence in a model trained with a
language modelling loss while concurrently minimizing the amount of contextual
information used to perform predictions. Finding ways of expressing such
constraints on the amount of information used during prediction, and verifying
the ways in which this can affect our results and LM pretraining in general
constitutes material for future work.

\begin{table}[t]
  \centering
  \rowcolors{2}{}{lightgray!20}
  \begin{tabular}{ll}
  \toprule
    connect-adj.\ baseline       & .42 \\
    \textbf{CPMI} (proj.) BERT \small{base multilingual cased}                  & .43 \\
  \midrule
    right-chain baseline & .40* \\
    \textbf{\fm{Dist} impact} (proj.) BERT \small{base uncased} & .52*  \\
  \bottomrule
  \small{*As reported in \citet[][Table 2]{wu.z:2021}}
  \end{tabular}
  \caption{UUAS on English PUD, for CPMI (from Table~\ref{tab:pud_uuas}), 
    compared to \citet{wu.z:2021}'s results. Note: the baselines aboves are
    theoretically identical, discrepancy may be due to data processing differences.}%
  \label{tab:impact-cpmi-comparison}
\end{table}










\input{acknowledgements.tex}

\bibliography{dependency-dependence}
\bibliographystyle{acl_natbib}

\clearpage
\pagebreak

\appendix
\input{appendix}

\input{appendix_additional-examples}

\end{document}

%% file: tikz-forest-parameters.tex
\usepackage[linguistics]{forest}
\forestset{%
    tt nonterminals/.style={
        for tree={%
            if={n_children==0}{}{font=\tt} }}}
\usepackage{tikz,tikz-dependency}
\pgfkeys{%
    /depgraph/edge unit distance=0.75ex,%
    /depgraph/reserved/edge style/.style = {%
        -, 
        semithick, solid, line cap=round, 
        rounded corners=2, 
    },%
    /depgraph/reserved/label style/.style = {%
    font=\ttfamily,
    scale = 0.5,
    text height = 1.5ex,
    text depth = 0.25ex, 
    inner sep = 0ex,%
    outer sep = 0pt,%
    text = black,%
    fill = white, 
    opacity = 0, text opacity = 0 
    }%
}

%% file: figures/tikz/ce/1352sum.projective_BERT-large-wgold.tikz
\begin{dependency}
	\begin{deptext}
		Gold\& was\& nowhere\& the\& spectacular\& performer\& it\& was\& two\& years\& ago\& on\& Black\& Monday\& . \\
	\end{deptext}
	\depedge{6}{1}{nsubj}
	\depedge{6}{2}{cop}
	\depedge{6}{3}{advmod}
	\depedge{6}{4}{det}
	\depedge{6}{5}{amod}
	\depedge{8}{7}{nsubj}
	\depedge{6}{8}{rcmod}
	\depedge{10}{9}{num}
	\depedge{11}{10}{npadvmod}
	\depedge{8}{11}{advmod}
	\depedge{8}{12}{prep}
	\depedge{14}{13}{nn}
	\depedge{12}{14}{pobj}
	\depedge[hide label, edge below, edge style={-, blue, opacity=0.5}]{10}{11}{}
	\depedge[hide label, edge below, edge style={-, blue, opacity=0.5}]{4}{6}{}
	\depedge[hide label, edge below, edge style={-, blue, opacity=0.5}]{5}{6}{}
	\depedge[hide label, edge below, edge style={-, blue, opacity=0.5}]{13}{14}{}
	\depedge[hide label, edge below, edge style={-, blue, opacity=0.5}]{9}{10}{}
	\depedge[hide label, edge below, edge style={-, blue, opacity=0.5}]{7}{8}{}
	\depedge[hide label, edge below, edge style={-, red, opacity=0.5}]{1}{2}{}
	\depedge[hide label, edge below, edge style={-, red, opacity=0.5}]{6}{7}{}
	\depedge[hide label, edge below, edge style={-, red, opacity=0.5}]{12}{13}{}
	\depedge[hide label, edge below, edge style={-, red, opacity=0.5}]{8}{9}{}
	\depedge[hide label, edge below, edge style={-, red, opacity=0.5}]{2}{3}{}
	\depedge[hide label, edge below, edge style={-, red, opacity=0.5}]{3}{4}{}
	\depedge[hide label, edge below, edge style={-, red, opacity=0.5}]{11}{12}{}
	\node (R) at (\matrixref.east) {{}};
	\node (R1) [right of = R] {\tiny\textsf{BERT-large}};
	\node (R4) at (R1.south) {\tiny $ 6/13 = 46\% $};
\end{dependency}

%% file: figures/tikz/ce/1352sum.projective_DistilBERT.tikz
\begin{dependency}
	\begin{deptext}
		Gold\& was\& nowhere\& the\& spectacular\& performer\& it\& was\& two\& years\& ago\& on\& Black\& Monday\& . \\
	\end{deptext}
	\depedge[hide label, edge below, edge style={-, blue, opacity=0.5}]{10}{11}{}
	\depedge[hide label, edge below, edge style={-, blue, opacity=0.5}]{12}{14}{}
	\depedge[hide label, edge below, edge style={-, blue, opacity=0.5}]{4}{6}{}
	\depedge[hide label, edge below, edge style={-, blue, opacity=0.5}]{5}{6}{}
	\depedge[hide label, edge below, edge style={-, blue, opacity=0.5}]{13}{14}{}
	\depedge[hide label, edge below, edge style={-, blue, opacity=0.5}]{9}{10}{}
	\depedge[hide label, edge below, edge style={-, blue, opacity=0.5}]{7}{8}{}
	\depedge[hide label, edge below, edge style={-, red, opacity=0.5}]{1}{2}{}
	\depedge[hide label, edge below, edge style={-, red, opacity=0.5}]{6}{7}{}
	\depedge[hide label, edge below, edge style={-, red, opacity=0.5}]{2}{3}{}
	\depedge[hide label, edge below, edge style={-, red, opacity=0.5}]{11}{14}{}
	\depedge[edge unit distance=0.55ex,
	hide label, edge below, edge style={-, red, opacity=0.5}]{7}{11}{}
	\depedge[hide label, edge below, edge style={-, red, opacity=0.5}]{3}{4}{}
	\node (R) at (\matrixref.east) {{}};
	\node (R1) [right of = R] {\tiny\textsf{DistilBERT}};
	\node (R4) at (R1.south) {\tiny $ 7/13 = 54\% $};
\end{dependency}

%% file: figures/tikz/ce/1352sum.projective_XLNet-base.tikz
\begin{dependency}
	\begin{deptext}
		Gold\& was\& nowhere\& the\& spectacular\& performer\& it\& was\& two\& years\& ago\& on\& Black\& Monday\& . \\
	\end{deptext}
	\depedge[hide label, edge below, edge style={-, blue, opacity=0.5}]{13}{14}{}
	\depedge[hide label, edge below, edge style={-, blue, opacity=0.5}]{12}{14}{}
	\depedge[hide label, edge below, edge style={-, blue, opacity=0.5}]{4}{6}{}
	\depedge[hide label, edge below, edge style={-, blue, opacity=0.5}]{9}{10}{}
	\depedge[hide label, edge below, edge style={-, red, opacity=0.5}]{1}{2}{}
	\depedge[hide label, edge below, edge style={-, red, opacity=0.5}]{6}{7}{}
	\depedge[hide label, edge below, edge style={-, red, opacity=0.5}]{4}{5}{}
	\depedge[edge unit distance=0.45ex,
	hide label, edge below, edge style={-, red, opacity=0.5}]{3}{8}{}
	\depedge[edge unit distance=0.55ex,
	hide label, edge below, edge style={-, red, opacity=0.5}]{10}{14}{}
	\depedge[hide label, edge below, edge style={-, red, opacity=0.5}]{2}{3}{}
	\depedge[hide label, edge below, edge style={-, red, opacity=0.5}]{3}{4}{}
	\depedge[hide label, edge below, edge style={-, red, opacity=0.5}]{11}{12}{}
	\depedge[edge unit distance=0.35ex,
	hide label, edge below, edge style={-, red, opacity=0.5}]{3}{14}{}
	\node (R) at (\matrixref.east) {{}};
	\node (R1) [right of = R] {\tiny\textsf{XLNet-base}};
	\node (R4) at (R1.south) {\tiny $ 4/13 = 31\% $};
\end{dependency}

%% file: acknowledgements.tex
\section*{Acknowledgements}

We thank anonymous reviewers, in particular the reviewer who alerted us to the
work of \citet{wu.z:2021}, and also Richard Futrell for helpful discussions and
feedback.
We also gratefully acknowledge support from the Centre for Research on Brain,
Language \& Music, the Natural Sciences and Engineering Research Council of
Canada, the Fonds de Recherche du Québec, Nature et Technologies and Société et
Culture, and the Canada CIFAR AI Chairs Program.

%% file: appendix.tex
\section{CPMI-dependency implementation details}

\subsection{Word2Vec as noncontextual PMI control}%
\label{sec:w2v}
We use Word2Vec \citep[][]{mikolov.t:2013} to obtain a non-conditional PMI
measure as a control/baseline. Additionally, in contrast with the CPMI values
extracted from contextual language models, this estimate does not take into
account the positions of the words in a particular sentence, but otherwise
reflects global distributional information similarly to the contextualized
models. Word2Vec should therefore function as a control with which to compare
the PMI estimates derived from the contextualized models.

Word2Vec maps a given word $w_i$ in the vocabulary it to a `target' embedding
vector $\mathbf{w}_i$, as well as an `context' embedding vector $\mathbf{c}_i$
(used during training). As demonstrated by \citet{levy.o:2014, allen.c:2019},
Word2Vec's training objective is optimized when the inner product of the target
and context embeddings equals the PMI, shifted by a global constant (determined
by $k$, the number of negative samples): \( \mathbf{w}_i^\top \mathbf{c}_j =
\pmi(w_i;w_j) -\log k. \) This type of embedding model thus provides a
non-contextual PMI estimator. A global shift will not change the resulting
PMI-dependency trees, so we simply take \(\pmi_\textrm{w2v}(w_i;w_j) \coloneqq
\mathbf{w}_i^\top \mathbf{c}_j \), with embeddings calculated using a Word2Vec
model trained on the same data as BERT.\footnote{We use the implementation in
\emph{Gensim} \citep[][]{rehurek.r:2010gensim}, trained on BookCorpus and
English Wikipedia, and use a global average vector for out-of-vocabulary words.}
Note: since we are ignoring the global shift of $k$, an absolute valued version
of PMI estimate will not be meaningful, and for this reason we only ever extract
dependencies from the Word2Vec PMI estimate without taking the absolute value.

\subsection{LtoR-CPMI for one-directional models}%
\label{sec:LtoR-CPMI} 
Our CPMI measure as defined above requires a bidirectional model (to calculate
probabilities of words given their context, both preceding and following). The
LSTM models we test in this study are left-to-right, so we define an slightly 
modified version of CPMI, to use with such unidirectional language models.
That is, for a left to right model $M_\mathrm{LR}$
\begin{align*}
&\CPMI_{M_\mathrm{LR}}({\bf w}_I;{\bf w}_J) = \\
&\log \frac{p_{M_\mathrm{LtoR}}({\bf w}_I\mid  {\bf w}_{0:I-1})}
           {p_{M_\mathrm{LtoR}}({\bf w}_I \mid {\bf w}_{0:J-1,J+1:I-1})},
\end{align*}
where ${\bf w}_{0:I-1}$ is the sentence up to before ${\bf w}_I$, and ${\bf
w}_{0:J-1,J+1:I-1}$ is the sentence up to before${\bf w}_I$, with ${\bf w}_J$
masked.

\subsection{Calculating CPMI scores}%
\label{sec:details-calculating-CPMI}

\subsubsection{Subtokenization}\label{sec:subtokenization}

We must formulate the CPMI measure between sequences of subtokens, rather than
tokens (words), because the large pretrained language models we use break down
words into subtokens, for which gold dependencies and part of speech tags are
not defined. 

The calculation of CPMI between two lists of subtokens ${\bf w}_I$ and ${\bf
w}_J$ in sentence ${\bf w}$ is
  \begin{align*}
    &\CPMI_{M}({\bf w}_I;{\bf w}_J) = \\
    &\log \frac{p_M({\bf w}_I\mid {\bf w}_{-I,J},{\bf w}_J)}
      {p_M({\bf w}_I\mid {\bf w}_{-I,J})}=
      \log \frac{p_M({\bf w}_I\mid {\bf w}_{-I})}
      {p_M({\bf w}_I\mid {\bf w}_{-I,J})}
  \end{align*}
where $I$ and $J$ are spans of (sub)token indices, ${\bf w}_I$ is the set of
subtokens with indices in $I$ (likewise for ${\bf w}_J$), ${\bf w}_{-I}$ is
the entire sentence without subtokens whose indices are in $I$, and ${\bf
w}_{-I,J}$ is the sentence without subtokens whose indices are in $I$ or $J$. 

Likewise, POS-CPMI is defined in terms of subtokens.  Note that gold POS tags
are defined for PTB word tokens, which may correspond to multiple subtokens.
POS-CPMI is calculated as:
  \begin{align*}
  &\mathrm{POS\text{-}CPMI}_M({\pi}_I;{\pi}_J) \\
  &= \log
  \frac{p_{M_\mathrm{POS}}({\pi}_I
        \mid  {\bf w}_{-I})}
        {p_{M_\mathrm{POS}}({\pi}_I
        \mid {\bf w}_{-I,J})}
  \end{align*}
where $M_\mathrm{POS}$ is the contextual embedding model $M$ with a POS
embedding network on top, and ${\pi}_I$ is the POS tag of ${\bf w}_I$ (the set
of subtokens with indices in $I$, as in the definition of CPMI above).

To get the probability estimate for a multiple-subtoken word, we use a
left-to-right chain rule decomposition. To get an estimate for a probability
$p({\bf w})$ of a subtokenized word ${\bf w}=w_0,w_1,\dots,w_n$ (that is, a
joint probability, which we cannot get straight from a language model), we use
a left-to-right chain rule decomposition of conditional probability estimates
within the word:
\[
  p({\bf w}) = p(w_0) \cdot p(w_1\mid w_0) \cdots p(w_n\mid w_{0:n-1})
\]
This decomposition allows us to estimate conditional pointwise information
between words made of multiple subtokens, at the expense of specifying a
left-to-right order within those words.


\subsubsection{Symmetrizing matrices}\label{sec:symmetrizing-matrices}
PMI is a symmetric function, but the estimated CPMI scores are not guaranteed to
be symmetric, since nothing in the models' training explicitly forces their
conditionaly probability estimates of words given context to respect the
identity $p(x | y)p(y)=p(y | x)p(x)$.
For this reason, we have a choice when assigning a score to a pair of words
$v,w$, whether we use the model's estimate of $\CPMI_{M}(v;w)$, which compares
the probability of $v$ with conditioner $w$ masked and unmasked, or of
$\CPMI_{M}(w;v)$. In our implementation of CPMI we calculate scores in both
directions, and use their sum (as mentioned in the main text
\S\ref{sec:extract-dependency-trees}), though experiments using one or the other
(using just the upper or lower triangular of the matrix), or the max (equivalent
to extracting a tree from the unsymmetrized matrix) led to very similar overall
results. Likewise for the Word2Vec PMI estimate, and the POS-CPMI estimates.

\subsubsection{Negative PMI values}%
\label{sec:absolute-valued-CPMI}
PMI may be positive or negative. Results in the main text are all computed for
CPMI dependencies extracted from signed matrices (so arcs with large negative
CPMI will be rarely included).  However, there is some discussion of
interpreting the magnitude of PMI as indicating dependency, independent of sign
\citep[see][]{salle.a:2019}. The choice to use an absolute-valued version of
CPMI might be justified by arguing that words which influence each other's
distribution should be connected, whether this influence is positive or
negative. 

In \S\ref{sec:results-supplemental-ptb} we include full results both with and without
taking the absolute value of the CPMI matrices before extracting trees. The
absolute-valued CPMI dependencies show a models increase in UUAS over the
corresponding matrices without taking the absolute value in general.  
But, it is not clear whether the choice to use absolute-valued CPMI would be
justified conceptually. Contrary to the conceptual motivation for CPMI
dependencies, in which words which often occur together should be linked, an
absolute-valued version links words which are highly informative of each others'
\emph{not} being present. For this reason we do not choose to use an
absolute-valued version of CPMI by default, but report those results for
comparison, note that the UUAS is in fact higher with the absolute value, and
refrain from further speculation.

\input{appendix_additional-CPMI.tex}

\section{Information Bottleneck for POS probe}%
\label{sec:IB-details}

The simple POS probe is a $d$-by-$h$-matrix, where the input dimension $h$ is
the contextual embedding network's hidden layer dimension, and the output
dimension $d$ is the number of different POS tags in the tagset. Interpreting
the output as an unnormalized probability distribution over POS tags, we train
the layer to minimize the cross-entropy loss between the predicted and observed
POS (using the labels from the Treebank). Training a simple linear probe is a
rough way to get a compressed representations from contextual embeddings, but it
has limitations \citep{hewitt.j:2019selectivity}.

A more correct way of extracting these representations is by a variational
information bottleneck technique \cite{tishby.n:2000}.  We implement this
technique \citep[roughly following][]{li.x:2019}, as follows. Optimization is to
minimize $\mathcal L_{\mathrm{IB}} =- I[Y;Z] + \beta I[H;Z]$, where $H$ is the
input embedding, $Z$ the latent representation and $Y$ the true label. This
technique trains two sets of parameters: the decoder, a linear model just as in
the simple linear POS probe, and the encoder, another linear model, whose output
in our case is interpreted as means and log-variances of a multivariate Gaussian
(a simplifying assumption). Minimizing this loss maximizes information in the
compressed representations about the output labels given a constraint on the
amount of information that the compressed representations carry about the
original embeddings.

\section{Equivalence of max pmi and max conditional probability objectives}%
\label{sec:hdpmi-hdcp}
\citet{maracek.d:2012} describes the equivalence of optimizing for trees with
maximum conditional probability of dependents given heads and optimizing for the
maximum PMI between dependents and heads. This equivalence relies on an
assumption that the marginal probability of words is independent of the parse
tree.

For a corpus $C$, a dependency structure $t$ can be described as a function
which maps the index of a word to the index of its head. If net mutual
information between dependents and heads according to dependency structure $t$
is $\pmi(t)\coloneqq \sum_{i} \pmi(w_i;w_{t(i)})$, and the log conditional
probability of dependents given heads is $\mathrm{\ell_{cond}}(t)\coloneqq
\prod_{w\in s} p(w_i\mid w_{t(i)})$, the optimum is the same:
\begin{align}
  &\mathop{\arg\max}_t \pmi(t) \\
  &= \mathop{\arg\max}_t \log \prod_{i=1}^{|C|} \frac{p(w_i,w_{t(i)})}{p(w_i)p(w_{t(i)})}\\
  \label{eq:maracek-step}
  &= \mathop{\arg\max}_t \log \prod_{i=1}^{|C|} \frac{p(w_i,w_{t(i)})}{p(w_{t(i)})}\\
  &= \mathop{\arg\max}_t \mathrm{\ell_{cond}}(t)
\end{align}
The step taken in~(\ref{eq:maracek-step}) follows only under the assumption that
the marginal probability of dependent words is independent of the structure $t$.
That is, that ``probabilities of the dependent words \ldots{} are the same for
all possible trees corresponding to a given sentence''~\citep[][\S
5.1.2]{maracek.d:2012}. This must be stipulated as an assumption in a
probabilistic model for the above derivation to hold.

\section{Augmented tables of results}%
\label{sec:additional-results}

We give results in further detail for the CPMI-dependencies on the English PTB
Wall Street Journal (WSJ) and on the multilingual PUD treebanks. Tables
described below follow this appendix.

\subsection{Results on WSJ data}%
\label{sec:results-supplemental-ptb}

Results presented in this section repeat those given in the main text, with two
independent additional parameters: projectivity and absolute value.

\paragraph{Projectivity} As described in \S\ref{sec:extract-dependency-trees},
in the main text we report results for projective CPMI dependency trees
extracted from CPMI matrices using Eisner's algorithm
\citet[][]{eisner.j:1996,eisner.j:1997}. These results are also repeated below,
but we additionally present UUAS results for maximum spanning trees (MSTs)
extracted from CPMI matrices using Prim's algorithm \citep{prim.r:1957},
following~\citet{hewitt.j:2019}.

\paragraph{Absolute value} In the main text we consider dependencies extracted
from signed CPMI matrices. As described in~\S\ref{sec:absolute-valued-CPMI}, we
also compute UUAS from absolute-valued matrices, and report them here.

\begin{itemize}
  \item Table~\ref{tab:cpmi-uuas-complete} is an augmented version of
  Table~\ref{tab:cpmi-uuas} from the main text, containing results for
  CPMI-dependencies both with and without the projectivity constraint.
  \item Table~\ref{tab:cpmi-uuas-complete-abs} is as the previous, but using an
  absolute valued version of CPMI.
  \item Table~\ref{tab:pos-cpmi-uuas-complete} is likewise an augmented version
  of Table~\ref{tab:pos-cpmi-uuas} from the main text, containing results for
  POS-CPMI-dependencies both with and without the projectivity constraint.
  \item Table~\ref{tab:pos-cpmi-uuas-complete-abs} is as the previous but using
  an absolute valued version of POS-CPMI.
\end{itemize}

In these tables, we also include the UUAS of randomized `lengthmatched' control.
For each sentence, this control consists of a randomized tree whose distribution
of arc lengths is identical to the gold tree (obtained by rejection sampling).

\subsubsection{WSJ10}%
\label{sec:wsj10}
Tables~\ref{tab:wsj10-cpmi-uuas-complete}
and~\ref{tab:wsj10-cpmi-uuas-complete-abs} give augmented UUAS results as in to
Tables~\ref{tab:cpmi-uuas-complete} and~\ref{tab:cpmi-uuas-complete-abs}, resp.,
but for only the sentences of length $\le 10$ from the test split (section 23)
of the WSJ corpus (WSJ10).  We include these results for comparison with much of
the unsupervised dependency parsing literature following
\citet{klein.d:2004induction}, which reports results on that subset. Note that
the UUAS is naturally higher across the board on this corpus of shorter
sentences.

\begin{table*}[t]
  \centering
  \rowcolors{9}{}{lightgray!20}
  \begin{tabular}{llcclcc}
  \toprule
                                   & \multicolumn{3}{c}{MSTs}&\multicolumn{3}{c}{Projective MSTs} \\
  \cmidrule(lr){2-4}\cmidrule(lr){5-7}
      {}                           & all & len $=1$ & len $>1$                                                & all & len $=1$ & len $>1$\\
      {}                           & {} & {\footnotesize prec. $|$ recall}& {\footnotesize prec. $|$ recall}  & {} & {\footnotesize prec. $|$ recall}& {\footnotesize prec. $|$ recall}\\
  \midrule
  random                           & .09 & .49 $|$ .10 & .05 $|$ .09                                          & .22 & .49 $|$ .34 & .08 $|$ .10\\
  connect-adjacent                 & \textbf{.49} & .49 $|$ 1\phantom{.0} & \phantom{.0}-- $|$ 0\phantom{.0}  & \textbf{.49} & .49 $|$ 1\phantom{.0} & \phantom{.0}-- $|$ 0\phantom{.0} \\
  lengthmatched                    & .37 &                     &                                            &  &                     &    \\
  \midrule
  \midrule
  Word2Vec                         & .27 & \textbf{.67} $|$ .36 & .13 $|$ .19                                 & .39 & \textbf{.61} $|$ .59 & .19 $|$ .19\\
  \midrule
  \textbf{BERT}  \small{base}      & .44 & .59 $|$ .68 & .26 $|$ .22                                          & .46 & .57 $|$ .72 & .27 $|$ .21\\
  \textbf{BERT}  \small{large}     & .46 & .56 $|$ \textbf{.79} & .23 $|$ .14                                 & .47 & .55 $|$ \textbf{.81} & .24 $|$ .13\\
  \textbf{DistilBERT}              & .46 & .58 $|$ .68 & \textbf{.30} $|$ .25                                 & .48 & .57 $|$ .72 & \textbf{.32} $|$ .24\\
  \textbf{Bart}  \small{large}     & .36 & .53 $|$ .60 & .15 $|$ .14                                          & .38 & .52 $|$ .64 & .16 $|$ .13\\
  \textbf{XLM}                     & .38 & .64 $|$ .55 & .20 $|$ .22                                          & .42 & .60 $|$ .64 & .23 $|$ .22\\
  \textbf{XLNet} \small{base}      & .42 & .61 $|$ .59 & .25 $|$ \textbf{.26}                                 & .45 & .59 $|$ .66 & .29 $|$ \textbf{.25}\\
  \textbf{XLNet} \small{large}     & .36 & .63 $|$ .51 & .19 $|$ .22                                          & .41 & .59 $|$ .61 & .23 $|$ .22\\
  \midrule
  \midrule
  vanilla LSTM                     & .40 & .56  $|$ .60 & .23 $|$ .22                                         & .44 & .54  $|$ .70 & .26 $|$ .19\\  
  \midrule
  \textbf{ONLSTM}                  & .41 & .57  $|$ .61 & .23 $|$ .22                                         & .44 & .55  $|$ .71 & .27 $|$ .19\\  
  \textbf{ONLSTM-SYD}              & .41 & .57  $|$ .61 & .23 $|$ .22                                         & .45 & .55  $|$ .71 & .27 $|$ .19\\
  \bottomrule
  \end{tabular}
  \caption{Total UUAS on the WSJ data, for CPMI dependencies extracted by both
    with a simple MST (Prim's algorithm; left) with a projectivity constraint
    (Eisner's algorithm; right, repeating Table~\ref{tab:cpmi-uuas}). In each
    case, overall scores are in the first column, followed by precision and
    recall UUAS for the subset consisting only of adjacent words (len $=$ 1),
    and likewise for nonadjacent words (len $>$ 1).}%
  \label{tab:cpmi-uuas-complete}
  \vspace*{20pt}
  \rowcolors{5}{}{lightgray!20}
  \begin{tabular}{llcclcc}
  \toprule
                                   & \multicolumn{3}{c}{MSTs}&\multicolumn{3}{c}{Projective MSTs} \\
  \cmidrule(lr){2-4}\cmidrule(lr){5-7}
      {}                           & all & len $=1$ & len $>1$                                                & all & len $=1$ & len $>1$\\
      {}                           & {} & {\footnotesize prec. $|$ recall}& {\footnotesize prec. $|$ recall}  & {} & {\footnotesize prec. $|$ recall}& {\footnotesize prec. $|$ recall}\\
  \midrule
  \textbf{BERT}  \small{base}      & .48 & .60 $|$ .75 & .29 $|$ .22                                          & .49 & .59 $|$ .78 & .31 $|$ .21\\
  \textbf{BERT}  \small{large}     & .48 & .56 $|$ \textbf{.84} & .25 $|$ .13                                 & .48 & .56 $|$ \textbf{.86} & .26 $|$ .13\\
  \textbf{DistilBERT}              & .48 & .58 $|$ .73 & \textbf{.32} $|$ .25                                 & .50 & .58 $|$ .77 & \textbf{.35} $|$ .24\\
  \textbf{Bart}  \small{large}     & .38 & .55 $|$ .59 & .19 $|$ .17                                          & .40 & .54 $|$ .64 & .20 $|$ .16\\
  \textbf{XLM}                     & .41 & .65 $|$ .59 & .22 $|$ .24                                          & .44 & \textbf{.63} $|$ .67 & .25 $|$ .23\\
  \textbf{XLNet} \small{base}      & .44 & .61 $|$ .62 & .27 $|$ \textbf{.26}                                 & .47 & .60 $|$ .70 & .30 $|$ \textbf{.25}\\
  \textbf{XLNet} \small{large}     & .37 & .63 $|$ .53 & .19 $|$ .23                                          & .42 & .61 $|$ .62 & .22 $|$ .22\\
  \midrule
  \midrule
  vanilla LSTM                     & .42 & .55  $|$ .63 & .25 $|$ .22                                         & .45 & .54  $|$ .73 & .28 $|$ .18\\  
  \midrule
  \textbf{ONLSTM}                  & .42 & .56  $|$ .63 & .25 $|$ .22                                         & .45 & .54  $|$ .73 & .29 $|$ .19\\  
  \textbf{ONLSTM-SYD}              & .42 & .56  $|$ .64 & .25 $|$ .22                                         & .46 & .54  $|$ .74 & .29 $|$ .19\\
  \bottomrule
  \end{tabular}
  \caption{As above in Table~\ref{tab:cpmi-uuas-complete}, but with dependencies
  extracted from absolute-valued matrices. As noted in \S\ref{sec:w2v}, due to
  the fact that Word2Vec estimates PMI only up to a global shift, an
  absolute-valued version would be meaningless, so we do not include that model
  here.}%
  \label{tab:cpmi-uuas-complete-abs}
\end{table*}

\begin{table*}[t]
  \centering
  \rowcolors{10}{}{lightgray!20}
  \begin{tabular}{lcccccc}
  \toprule
                                   & \multicolumn{3}{c}{MSTs}&\multicolumn{3}{c}{Projective MSTs} \\
  \cmidrule(lr){2-4}\cmidrule(lr){5-7}    
      {}                           & all & len $=1$ & len $>1$                                                & all & len $=1$ & len $>1$                                                \\
      {}                           & {} & {\footnotesize prec. $|$ recall}& {\footnotesize prec. $|$ recall}      & {} & {\footnotesize prec. $|$ recall}& {\footnotesize prec. $|$ recall}      \\
  \toprule
  random                           & .29 & .56 $|$ .30 & .18 $|$ .28                                          & .34 & .54 $|$ .45 & .18 $|$ .21                                          \\
  adjacent                         & \textbf{.53} & .53 $|$ 1\phantom{.0}  & \phantom{.0}-- $|$ 0\phantom{.0} & \textbf{.53} & .53 $|$ 1\phantom{.0}  & \phantom{.0}-- $|$ 0\phantom{.0} \\
  lengthmatched                    & .51                                                                      &                                                                       \\
  \midrule
  \midrule
  Word2Vec                         & .42 & \textbf{.61} $|$ .51 & .28 $|$ .32                                 & .46 & \textbf{.60} $|$ .63 & .29 $|$ .27                                 \\ 
  \midrule
  \textbf{BERT}  \small{base}      & .51 & .60 $|$ .69 & .36 $|$ .29                                          & .52 & .59 $|$ .72 & .38 $|$ .28                                          \\ 
  \textbf{BERT}  \small{large}     & .52 & .59 $|$ \textbf{.81} & .34 $|$ .20                                 & \textbf{.53} & .59 $|$ \textbf{.82} & .36 $|$ .20                                 \\ 
  \textbf{DistilBERT}              & .51 & .59 $|$ .71 & \textbf{.38} $|$ .29                                 & .52 & .58 $|$ .75 & \textbf{.40} $|$ .27                                 \\ 
  \textbf{Bart}  \small{large}     & .44 & .54 $|$ .63 & .27 $|$ .21                                          & .45 & .54 $|$ .66 & .28 $|$ .21                                          \\ 
  \textbf{XLM}                     & .48 & \textbf{.61} $|$ .61 & .32 $|$ .32               & .49 & \textbf{.60} $|$ .66 & \textbf{.34} $|$ \textbf{.31}               \\ 
  \textbf{XLNet} \small{base}      & .51 & \textbf{.61} $|$ .64 & \textbf{.38} $|$ \textbf{.35}                                 & \textbf{.53} & \textbf{.60} $|$ .69 & .42 $|$ .35                                 \\ 
  \textbf{XLNet} \small{large}     & .46 & \textbf{.61} $|$ .57 & .32 $|$ .34                                 & .48 & .59 $|$ .64 & .34 $|$ .31                                 \\ 
  \bottomrule
  \end{tabular}
  \caption{Total UUAS on WSJ10, for CPMI dependencies extracted both
    without the projectivity constraint (MSTs), and with it (Projective MSTs).
    Compare with an overall UUAS of \textbf{.637} reported in
    \citet[][Fig.~3]{klein.d:2004induction} for the complete WSJ10.}%
  \label{tab:wsj10-cpmi-uuas-complete}
  \vspace*{20pt}
  \rowcolors{5}{}{lightgray!20}
  \begin{tabular}{lcccccc}
  \toprule
                                   & \multicolumn{3}{c}{MSTs}&\multicolumn{3}{c}{Projective MSTs} \\
  \cmidrule(lr){2-4}\cmidrule(lr){5-7}
      {}                           & all & len $=1$ & len $>1$                                                & all & len $=1$ & len $>1$                                                \\
      {}                           & {} & {\footnotesize prec. $|$ recall}& {\footnotesize prec. $|$ recall}      & {} & {\footnotesize prec. $|$ recall}& {\footnotesize prec. $|$ recall}      \\
  \midrule
  \textbf{BERT}  \small{base}      & .53 & .60 $|$ .75 & .39 $|$ .28                                          & .54 & .60 $|$ .78 & .41 $|$ .27                                          \\ 
  \textbf{BERT}  \small{large}     & .54 & .60 $|$ \textbf{.85} & .37 $|$ .19                                 & .54 & .59 $|$ \textbf{.86} & .38 $|$ .19                                 \\ 
  \textbf{DistilBERT}              & .54 & .60 $|$ .77 & \textbf{.41} $|$ .28                                 & .55 & .60 $|$ .79 & \textbf{.43} $|$ .27                                 \\ 
  \textbf{Bart}  \small{large}     & .47 & .58 $|$ .63 & .31 $|$ .28                                          & .48 & .58 $|$ .67 & .33 $|$ .27                                          \\ 
  \textbf{XLM}                     & .50 & \textbf{.64} $|$ .65 & \textbf{.33} $|$ \textbf{.32}               & .51 & \textbf{.63} $|$ .69 & .35 $|$ \textbf{.31}               \\ 
  \textbf{XLNet} \small{base}      & .52 & \textbf{.62} $|$ .68 & .39 $|$ .34                                 & .55 & .62 $|$ .73 & .42 $|$ .34                                 \\ 
  \textbf{XLNet} \small{large}     & .48 & \textbf{.62} $|$ .61 & .33 $|$ .34                                 & .51 & .61 $|$ .66 & .37 $|$ .33                                 \\ 
  \bottomrule
  \end{tabular}
  \caption{Total UUAS on WSJ10, MST and Projective MST, as above, but extracted from absolute-valued CPMI matrices.}%
  \label{tab:wsj10-cpmi-uuas-complete-abs}
\end{table*}

\begin{table*}
  \centering
  \rowcolors{5}{}{lightgray!20}
  \begin{tabular}{rllcclcc} 
  \toprule
    &         & \multicolumn{3}{c}{MSTs}&\multicolumn{3}{c}{Projective MSTs} \\
  \cmidrule(lr){3-5}\cmidrule(lr){6-8}
  {}&         & all & len $= 1$ & len $> 1$                                         & all & len $= 1$ & len $> 1$  \\
  {}&         & {} & {\footnotesize prec. $|$ recall}& {\footnotesize prec. $|$ recall} & {}  & {\footnotesize prec. $|$ recall}& {\footnotesize prec. $|$ recall}\\
  \midrule 
  \cellcolor{white}&\textbf{BERT}  \small{base}   & .47 & .57 $|$ .77& .29 $|$ .20  & .48 & .56 $|$ .79 & .32 $|$ .19\\
  \cellcolor{white}&\textbf{BERT}  \small{large}  & .44 & .54 $|$ .73& .25 $|$ .17  & .45 & .53 $|$ .75 & .27 $|$ .16\\
  \cellcolor{white}&\textbf{XLNet} \small{base}   & .29 & .56 $|$ .41& .14 $|$ .17  & .36 & .55 $|$ .56 & .17 $|$ .17\\
  \multirow{-4}{*}{\rotatebox[origin=c]{90}{simple-POS}}
  \cellcolor{white}&\textbf{XLNet} \small{large}  & .26 & .59 $|$ .38& .11 $|$ .15  & .32 & .56 $|$ .51 & .14 $|$ .15\\
  \midrule
  \cellcolor{white}&\textbf{BERT}  \small{base}   & .38 & .60 $|$ .58 & .18 $|$ .18 & .41 & .58 $|$ .65 & .20 $|$ .18\\
  \cellcolor{white}&\textbf{BERT}  \small{large}  & .39 & .56 $|$ .64 & .17 $|$ .14 & .41 & .55 $|$ .69 & .18 $|$ .14\\
  \cellcolor{white}&\textbf{XLNet} \small{base}   & .36 & .57 $|$ .52 & .19 $|$ .20 & .40 & .55 $|$ .60 & .22 $|$ .20\\
  \multirow{-4}{*}{\rotatebox[origin=c]{90}{IB-POS}}
  \cellcolor{white}&\textbf{XLNet} \small{large}  & .30 & .60 $|$ .44 & .13 $|$ .17 & .36 & .56 $|$ .56 & .16 $|$ .16\\
  \bottomrule
  \end{tabular}
  \caption{Total UUAS for POS-CPMI, both MST (left) and projective MST (right, a repeat of Table~\ref{tab:pos-cpmi-uuas}),
  using the simple POS probe and IB POS probe, from BERT and XLNet models.}%
  \label{tab:pos-cpmi-uuas-complete}
  \vspace*{20pt}
  \rowcolors{5}{}{lightgray!20}
  \begin{tabular}{rllcclcc} 
  \toprule
    &         & \multicolumn{3}{c}{MSTs}&\multicolumn{3}{c}{Projective MSTs} \\
  \cmidrule(lr){3-5}\cmidrule(lr){6-8}
  {}&         & all & len $= 1$ & len $> 1$                                         & all & len $= 1$ & len $> 1$  \\
  {}&         & {} & {\footnotesize prec. $|$ recall}& {\footnotesize prec. $|$ recall} & {}  & {\footnotesize prec. $|$ recall}& {\footnotesize prec. $|$ recall}\\
  \midrule 
  \cellcolor{white}&\textbf{BERT}  \small{base}   & .49 & .57 $|$ .78& .32 $|$ .21  & .50 & .57 $|$ .80 & .34 $|$ .21\\
  \cellcolor{white}&\textbf{BERT}  \small{large}  & .47 & .56 $|$ .79& .28 $|$ .17  & .48 & .55 $|$ .81 & .30 $|$ .16\\
  \cellcolor{white}&\textbf{XLNet} \small{base}   & .31 & .57 $|$ .44& .15 $|$ .18  & .36 & .56 $|$ .56 & .17 $|$ .17\\
  \multirow{-4}{*}{\rotatebox[origin=c]{90}{simple-POS}}
  \cellcolor{white}&\textbf{XLNet} \small{large}  & .27 & .59 $|$ .40& .12 $|$ .15  & .31 & .57 $|$ .49 & .13 $|$ .14\\
  \midrule
  \cellcolor{white}&\textbf{BERT}  \small{base}   & .35 & .60 $|$ .52 & .16 $|$ .18 & .39 & .59 $|$ .61 & .19 $|$ .18\\
  \cellcolor{white}&\textbf{BERT}  \small{large}  & .40 & .58 $|$ .67 & .17 $|$ .15 & .43 & .57 $|$ .72 & .19 $|$ .14\\
  \cellcolor{white}&\textbf{XLNet} \small{base}   & .38 & .58 $|$ .56 & .20 $|$ .21 & .42 & .57 $|$ .63 & .23 $|$ .21\\
  \multirow{-4}{*}{\rotatebox[origin=c]{90}{IB-POS}}
  \cellcolor{white}&\textbf{XLNet} \small{large}  & .30 & .59 $|$ .44 & .13 $|$ .16 & .35 & .57 $|$ .55 & .16 $|$ .16\\
  \bottomrule
  \end{tabular}
  \caption{As above in Table~\ref{tab:pos-cpmi-uuas-complete}, but with
  dependencies extracted from absolute-valued matrices.}%
  \label{tab:pos-cpmi-uuas-complete-abs}
\end{table*}

\subsection{Results on multilingual PUD data}%
\begin{figure*}
  \small
  \centering
  \rowcolors{5}{lightgray!20}{}
  \begin{tabular}{llrrrrrrr}
    \toprule
    {}         &              & \multicolumn{7}{c}{UUAS}\\
    \cmidrule(lr){3-9}
    {}         &              &         &  \multicolumn{3}{c}{MSTs}       & \multicolumn{3}{c}{Projective MSTs} \\
    \cmidrule(lr){4-6}\cmidrule(lr){7-9}
    language   & mean sent.\ length & connect-adjacent &  random  &  CPMI &  CPMI (abs)  & random &  CPMI &  CPMI (abs) \\
    \midrule
    Arabic     &        17.52 &     .58 &      .11 &   .43 &          .48 &    .27 &   .45 &         .51 \\
    Chinese    &        17.51 &     .45 &      .11 &   .38 &          .39 &    .23 &   .40 &         .42 \\
    Czech      &        14.99 &     .48 &      .12 &   .47 &          .48 &    .25 &   .48 &         .50 \\
    English    &        17.73 &     .42 &      .10 &   .41 &          .43 &    .22 &   .43 &         .45 \\
    Finnish    &        12.47 &     .52 &      .15 &   .45 &          .46 &    .28 &   .47 &         .48 \\
    French     &        21.18 &     .45 &      .08 &   .44 &          .46 &    .23 &   .47 &         .49 \\
    German     &        17.56 &     .42 &      .11 &   .44 &          .46 &    .22 &   .46 &         .48 \\
    Hindi      &        20.53 &     .51 &      .09 &   .38 &          .39 &    .24 &   .41 &         .42 \\
    Icelandic  &        15.88 &     .49 &      .12 &   .40 &          .41 &    .25 &   .42 &         .44 \\
    Indonesian &        16.06 &     .56 &      .12 &   .44 &          .46 &    .27 &   .46 &         .49 \\
    Italian    &        20.43 &     .45 &      .09 &   .45 &          .46 &    .23 &   .47 &         .48 \\
    Japanese   &        24.73 &     .48 &      .08 &   .30 &          .39 &    .23 &   .34 &         .43 \\
    Korean     &        13.99 &     .58 &      .13 &   .46 &          .48 &    .28 &   .49 &         .50 \\
    Polish     &        14.73 &     .54 &      .12 &   .50 &          .51 &    .27 &   .52 &         .53 \\
    Portuguese &        19.83 &     .45 &      .10 &   .44 &          .46 &    .23 &   .47 &         .48 \\
    Russian    &        15.38 &     .51 &      .12 &   .49 &          .50 &    .26 &   .51 &         .51 \\
    Spanish    &        20.00 &     .45 &      .09 &   .46 &          .47 &    .23 &   .48 &         .50 \\
    Swedish    &        16.14 &     .44 &      .11 &   .41 &          .43 &    .24 &   .43 &         .45 \\
    Thai       &        21.05 &     .56 &      .09 &   .39 &          .38 &    .25 &   .42 &         .42 \\
    Turkish    &        13.73 &     .55 &      .14 &   .46 &          .48 &    .27 &   .48 &         .50 \\
    \bottomrule
  \end{tabular}
  \caption{UUAS for multilingual Parallel UD dataset, for CPMI
  dependencies extracted from from BERT base multilingual. Note that while the
  dataset consists of the same 1000 sentences translated into the 20
  languages, there is some variation across languages in mean sentence length.
  Projective (signed) UUAS are plotted below in
  Figure~\ref{fig:pud_uuas-complete} with random and connect-adjacent
  baselines.}%
  \label{tab:pud_uuas-complete}
  \vspace*{20pt}
  \includegraphics[width=\linewidth]{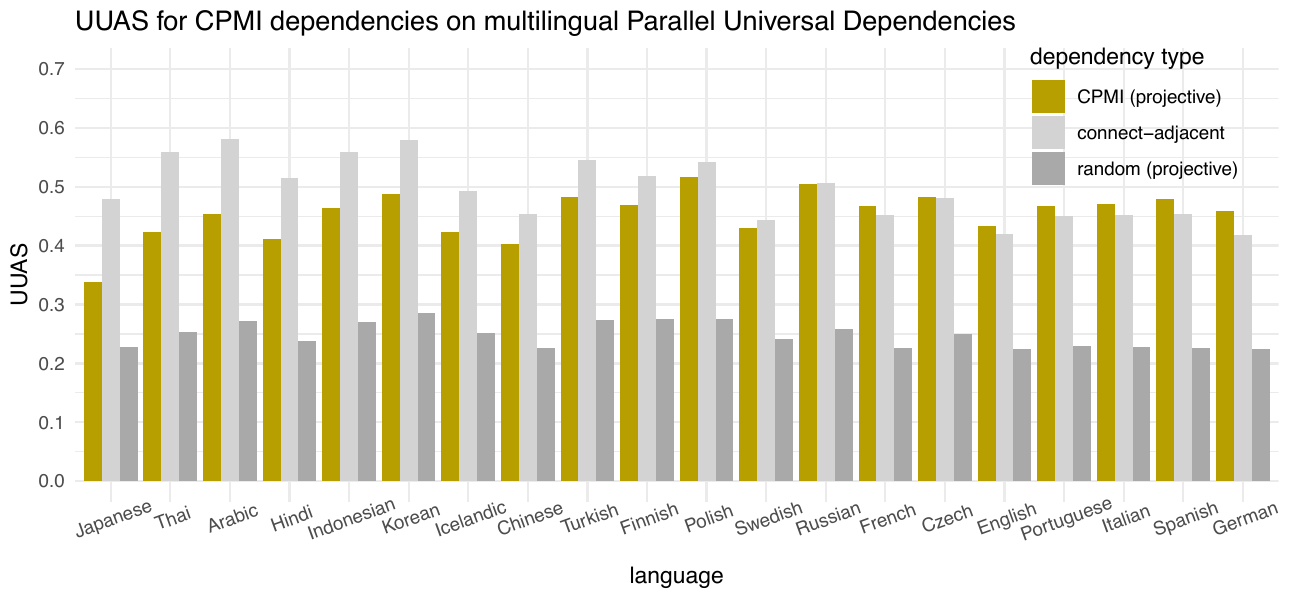}
  \caption{CPMI UUAS (signed, projective) from BERT base multilingual, ordered
  by the difference between CPMI UUAS and the connect-adjacent baseline UUAS.
  For most languages the CPMI UUAS is below or comparable to the
  connect-adjacent baseline.}%
  \label{fig:pud_uuas-complete}
\end{figure*}

\label{sec:results-supplemental-pud}
Table~\ref{tab:pud_uuas-complete} gives results on the 20 languages of the
Parallel Universal Dependencies (PUD) treebanks.  
These parallel treebanks were included in the CoNLL 2017 shared task on
Multilingual Parsing from Raw Text to Universal Dependencies. The PUD treebank
for each language consists of 1000 sentences annotated for Universal
Dependencies. The sentences are translated into each of the languages, with the
majority (750) being originally in English.

We compute CPMI for these sentences using the multilingual pretrained BERT-base
model made available by Hugging Face Transformers
\citep{wolf.t:2020transformers}.\footnote{\url{https://huggingface.co/bert-base-multilingual-cased}}
This model was trained using masked language modelling and next sentence
prediction on the 104 languages with the largest Wikipedias, including all 20 in
the PUD. UUAS for CPMI dependency trees for all languages is plotted in
Figure~\ref{fig:pud_uuas-complete}.

%% file: appendix_additional-CPMI.tex



\subsection{Additional analysis of CPMI dependencies}\label{sec:cpmi-additional-analyses}

\subsubsection{Similarity between models}\label{sec:model-similarity}
\begin{figure}
  \centering \includegraphics[%
  trim={0 2cm 0 .3cm}, clip, width=\linewidth%
  ]{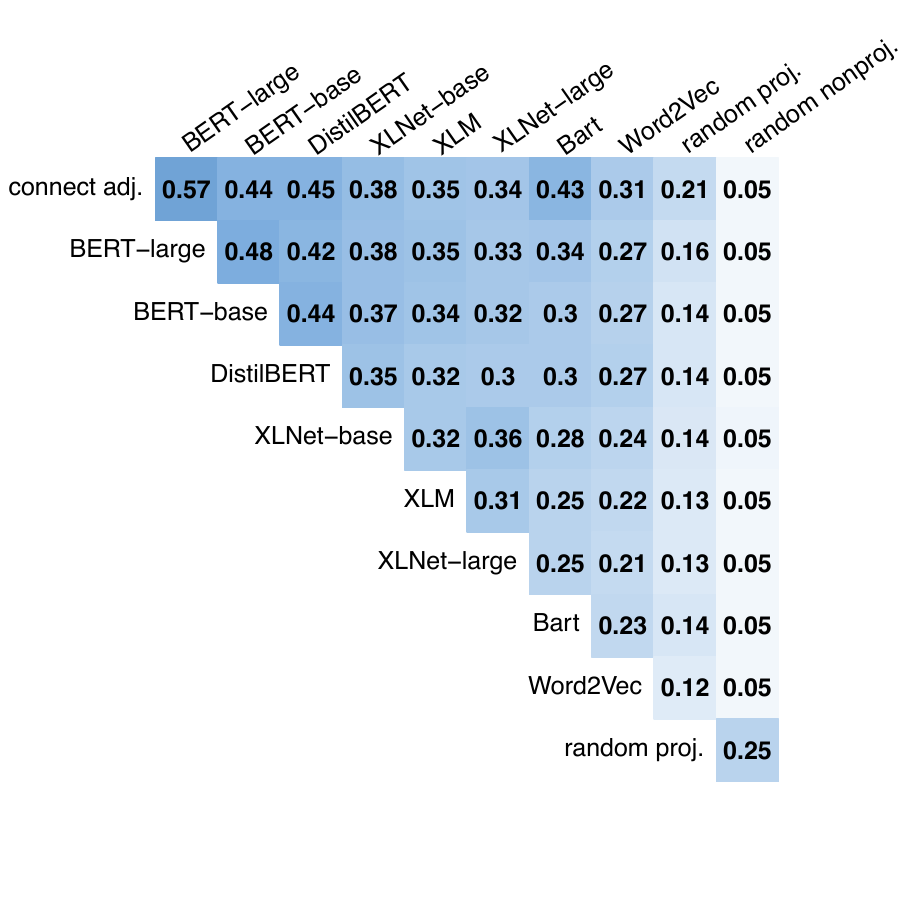}
  \caption{Similarity of models' predictions, by wordpair, reported as
    Jaccard index, the intersection of the two models' sets of dependency
  edges divided their union.}%
  \label{fig:model-similarity}
\end{figure}

Figure~\ref{fig:model-similarity} shows the similarity of the CPMI dependency
structures extracted from the different contextual embedding models.
We measure similarity of dependency structures with the
Jaccard index for the sets of the predicted edges by two models.
Jaccard index measures similarity of two sets $A,B$ and is defined as $J(A,B)
= |A\cap B|/|A\cup B|$. The contextualized models agree with each other on around 30--50\% of the edges, and agree with the the noncontextual baseline W2V slightly less. In general, they agree with the linear baseline at somewhat higher rates.

\subsubsection{Accuracy versus arc length}%
\label{sec:acc-vs-len}

\begin{figure}
  \centering
  \includegraphics[width=\linewidth]{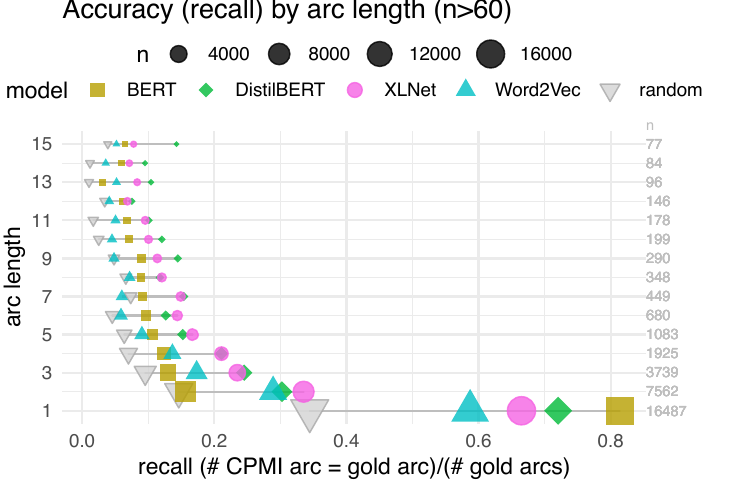}
  \caption{Recall accuracy is higher for shorter arcs.  The distinction is
    mostly between arcs of length 1 vs longer arcs.
    Note that the relatively higher accuracy of BERT (large)'s estimates overall
  are driven by its very large proportion of length 1 arcs.}%
  \label{fig:lindist}
\end{figure}

Breaking down the results by dependency length,
Figure~\ref{fig:lindist}
shows the recall accuracy of CPMI dependencies, grouped by length of gold arc.
In general, length 1 arcs have the highest accuracy;
longer dependencies have lower accuracy.
CPMI dependencies from BERT (large) have 81\% recall accuracy on length 1 arcs,
with arcs longer than 1 having much lower recall (13\% overall) near random (10\%).
In other models, XLNet in particular, this distinction is less of a binary distinction,
but the trend is still for lower recall on longer arcs.

\subsubsection{Accuracy versus perplexity}%
\label{sec:acc-vs-ppl}

Here we investigate the correlation between language model performance and
CPMI-dependency accuracy. If models' confidence in predicting were tied to
accuracy, it would be hard to argue that the relatively low accuracy score we
see was due to the lack of connection between syntactic dependency and
statistical dependency, rather than to the models' struggling to recover such
a structure. Here we measure model confidence by obtaining a perplexity score
for each sentence, calculated as the negative mean of the pseudo
log-likelihood, that is, for a sentence $\mathbf w$ of length $n$,
\[\mathrm{pseudo~PPL}(\mathbf w)=\exp{[-\frac{1}{n}\sum_{I=1}^n \log p(\mathbf{w}_I|\mathbf{w}_{-I})]}\]

\begin{figure}
  \centering
  \includegraphics[width=\linewidth]{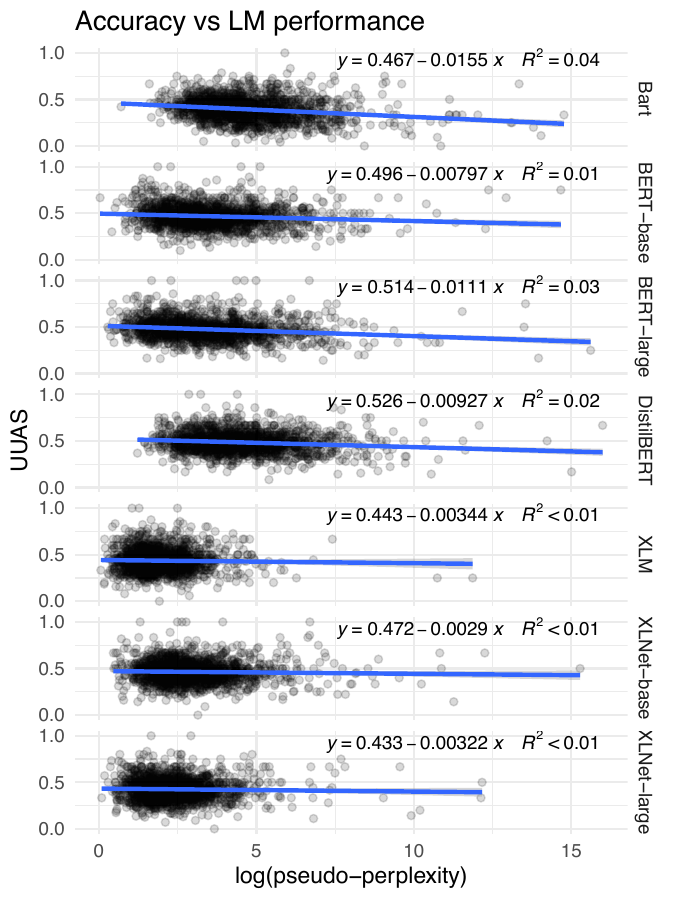}
  \caption{Per-sentence accuracy (UUAS) against log psuedo-perplexity. Accuracy
    is not tied to the confidence of the language model on a given sentence, for
    any of the models (there is a slight tendency to have higher accuracy on
    sentences of lower perplexity, but the effect size is negligible, and
  correlation is very low).}%
  \label{fig:uas-vs-ppl}
\end{figure}

Figure~\ref{fig:uas-vs-ppl} shows that accuracy is not correlated with
sentence-level perplexity for any of the models (fitting a linear regression,
$R^2<0.05$ for each model). That is, the accuracy of CPMI-dependency
structures is roughly the same on the sentences which the model predicts
confidently (lower perplexity) as on the sentences which it predicts less
confidently (higher perplexity).

\subsubsection{UUAS during training}%
\begin{figure}
  \includegraphics[width=\linewidth]{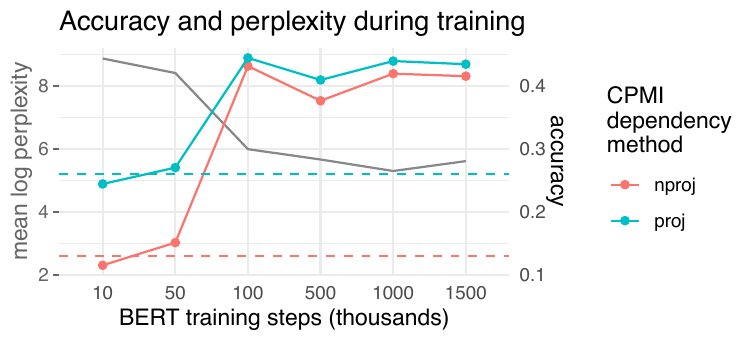}
  \caption{Training checkpoints for BERT-base uncased. After about 1 million
    training steps, the perplexity (gray, axis left) has plateaued. The UUAS
    (axis right) of extracted CPMI structures does not increase past the level
  it reaches at 100k steps.}%
  \label{fig:checkpoints}
\end{figure}

\label{sec:checkpoints}
We examined the accuracy of CPMI dependencies during training of BERT (base uncased) from scratch.
Figure~\ref{fig:checkpoints} shows the average perplexity of this model, along
with the sentence-wise average accuracy of CPMI structures at selected
checkpoints during training. After about one million training steps
the model has reached a plateau in terms of performance (perplexity stops
decreasing), and we see that the peak UUAS has also plateaued at that point,
but in fact reached its highest value after one hundred thousand training steps.


\subsubsection{UUAS by dependency label}
Table~\ref{tab:uuas-label-comparison} gives per-dependency label recall accuracy of CPMI-dependencies extracted from  the subset of dependency labels for which XLNet (base) achieves UUAS higher than both the linear and a random (projective) baselines.

\newcolumntype{L}{>{\sffamily}l}
\begin{table*}[t]
  \footnotesize
  \centering
  \begin{tabular}{L|ll|rrrrrr|rr}
    \toprule
    \textrm{relation} & meanlen  &   n  & BERT & DistilBERT & Bart & XLNet &  XLM &  W2V & connect adj. & rand proj. \\
    \midrule
    xcomp     &     3.1  &  398 & 0.24 &       0.23 & 0.18 &  \textbf{0.43} & 0.40 & 0.26 &    0.07 &   0.13   \\
    mark      &     5.0  &  421 & 0.18 &       0.29 & 0.11 &  \textbf{0.30} & 0.20 & 0.09 &    0.05 &   0.10   \\
    conj      &     6.1  & 1009 & 0.12 &       0.19 & 0.21 &  0.28 & 0.26 & \textbf{0.29} &    0.03 &   0.10   \\
    ccomp     &     6.9  &  550 & 0.11 &       0.15 & 0.07 &  \textbf{0.19} & 0.14 & 0.06 &    0.03 &   0.08   \\
    dobj      &     2.4  & 1637 & 0.37 &       0.38 & 0.33 &  \textbf{0.47} & 0.42 & 0.35 &    0.21 &   0.16   \\
    advcl     &     8.7  &  293 & 0.05 &       0.04 & 0.05 &  \textbf{0.11} & 0.07 & 0.06 &    0.00 &   0.06   \\
    nsubjpass &     4.3  &  253 & 0.13 &       0.15 & 0.12 &  0.21 & \textbf{0.26} & 0.19 &    0.00 &   0.13   \\
    rcmod     &     4.1  &  290 & 0.11 &       0.07 & 0.12 &  0.12 & \textbf{0.14} & 0.11 &    0.00 &   0.08   \\
    poss      &     2.4  &  709 & 0.30 &       0.28 & 0.21 &  \textbf{0.32} & 0.31 & 0.30 &    0.24 &   0.17   \\
    pobj      &     2.3  & 3745 & 0.33 &       \textbf{0.39} & 0.28 &  0.36 & 0.32 & 0.30 &    0.30 &   0.17   \\
    tmod      &     3.0  &  244 & 0.31 &       0.35 & 0.30 &  0.39 & \textbf{0.40} & 0.18 &    0.33 &   0.18   \\
    cop       &     2.1  &  330 & 0.39 &       \textbf{0.49} & 0.39 &  0.42 & 0.33 & 0.33 &    0.39 &   0.22   \\
    det       &     1.7  & 3327 & 0.52 &       \textbf{0.64} & 0.24 &  0.53 & 0.43 & 0.41 &    0.52 &   0.23   \\
    \bottomrule
  \end{tabular}
  \caption{Recall accuracy by label for the labels which XLNet achieves above the baselines, for the models BERT large, Distilbert base, Bart large, XLNet base, XLM, as well as Word2Vec, and the connect adjacent and random baselines.}
  \label{tab:uuas-label-comparison}
\end{table*}

%% file: appendix_additional-examples.tex
  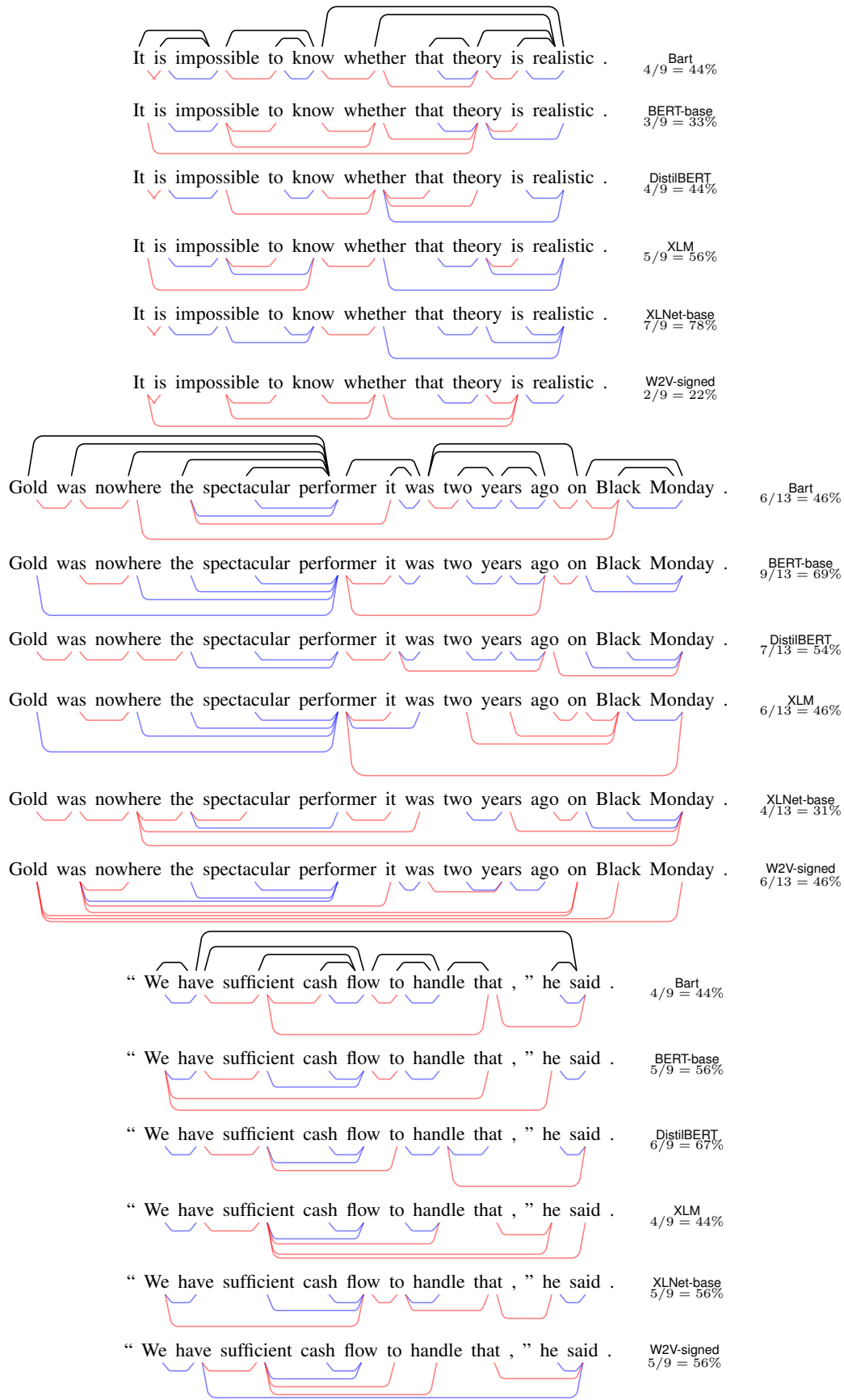
\begin{figure*}[t]
    \begin{footnotesize}
      \centering
      \input{figures/tikz/ce/442sum.projective_Bart-wgold.tikz}\\
      \input{figures/tikz/ce/442sum.projective_BERT-base.tikz}\\
      \input{figures/tikz/ce/442sum.projective_DistilBERT.tikz}\\
      \input{figures/tikz/ce/442sum.projective_XLM.tikz}\\
      \input{figures/tikz/ce/442sum.projective_XLNet-base.tikz}\\
      \input{figures/tikz/ce/442sum.projective_W2V-signed.tikz}\\

      \input{figures/tikz/ce/1352sum.projective_Bart-wgold.tikz}\\
      \input{figures/tikz/ce/1352sum.projective_BERT-base.tikz}\\
      \input{figures/tikz/ce/1352sum.projective_DistilBERT.tikz}\\
      \input{figures/tikz/ce/1352sum.projective_XLM.tikz}\\
      \input{figures/tikz/ce/1352sum.projective_XLNet-base.tikz}\\
      \input{figures/tikz/ce/1352sum.projective_W2V-signed.tikz}\\

      \input{figures/tikz/ce/405sum.projective_Bart-wgold.tikz}\\
      \input{figures/tikz/ce/405sum.projective_BERT-base.tikz}\\
      \input{figures/tikz/ce/405sum.projective_DistilBERT.tikz}\\
      \input{figures/tikz/ce/405sum.projective_XLM.tikz}\\
      \input{figures/tikz/ce/405sum.projective_XLNet-base.tikz}\\
      \input{figures/tikz/ce/405sum.projective_W2V-signed.tikz}
      \caption{Additional examples of projective parses from Bart, BERT,
        DistilBERT, XLM, XLNet, and the noncontextual baseline Word2Vec. Gold
        standard dependency parse above in black, CPMI-dependencies below, blue
        where they agree with gold dependencies, and red when they do not.
      Accuracy scores (UUAS) are given for each sentence.}%
      \label{fig:further-examples}
    \end{footnotesize}
  \end{figure*}



  \begin{figure*}[t]
    \centering \includegraphics[width=\linewidth]{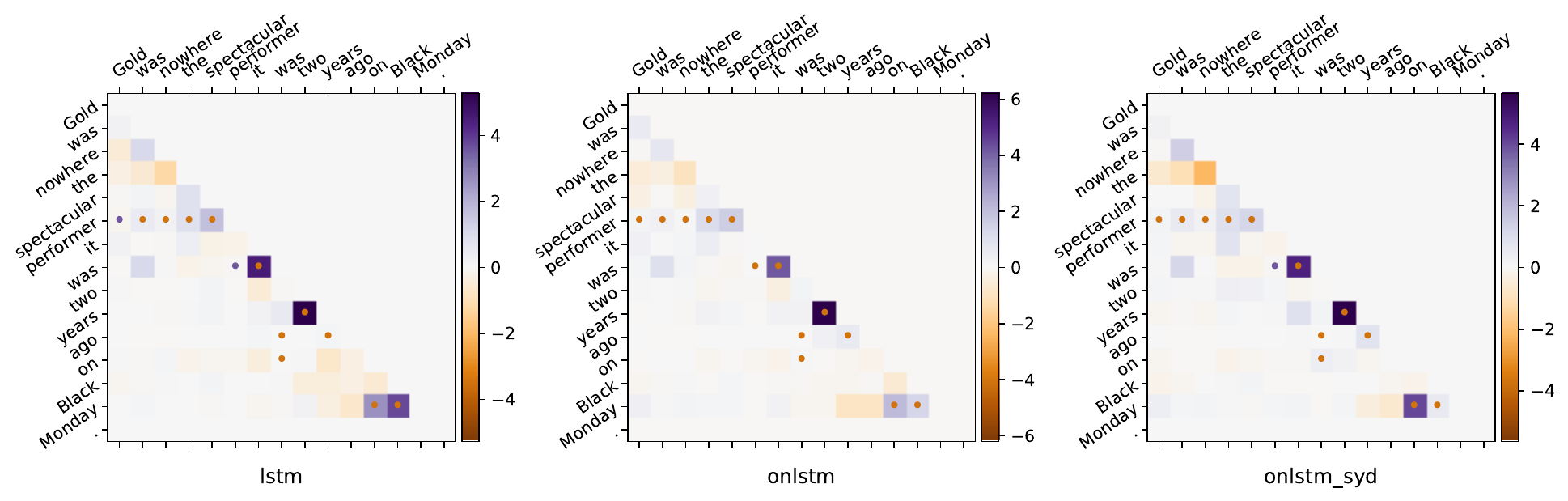}
    \caption{CPMI matrices for ONLSTM and ONLSTM-SYD, with vanilla LSTM
      baseline. Gold edges are marked with a dot. Compare with
    dependency structures in Figure~\ref{fig:LSTM-examples}}%
    \label{fig:example-matrices-lstms}
    \begin{footnotesize}
      \centering
      \input{figures/tikz/lstm/442sum.projective_LSTM-wgold.tikz}\\
      \input{figures/tikz/lstm/442sum.projective_ONLSTM.tikz}\\
      \input{figures/tikz/lstm/442sum.projective_ONLSTM-SYD.tikz}\\

      \input{figures/tikz/lstm/1352sum.projective_LSTM-wgold.tikz}\\
      \input{figures/tikz/lstm/1352sum.projective_ONLSTM.tikz}\\
      \input{figures/tikz/lstm/1352sum.projective_ONLSTM-SYD.tikz}\\

      \input{figures/tikz/lstm/405sum.projective_LSTM-wgold.tikz}\\
      \input{figures/tikz/lstm/405sum.projective_ONLSTM.tikz}\\
      \input{figures/tikz/lstm/405sum.projective_ONLSTM-SYD.tikz}
      \caption{Projective parses from the LSTM baseline and the ONSLTM and
        syntactic (ONSLTM-SYD) models for three example sentences. Matrices for
      the second sentence are in Figure~\ref{fig:example-matrices-lstms}.}%
      \label{fig:LSTM-examples}
    \end{footnotesize}
  \end{figure*}

%% file: figures/tikz/ce/442sum.projective_Bart-wgold.tikz
\begin{dependency}
	\begin{deptext}
		It\& is\& impossible\& to\& know\& whether\& that\& theory\& is\& realistic\& . \\
	\end{deptext}
	\depedge{3}{1}{nsubj}
	\depedge{3}{2}{cop}
	\depedge{5}{4}{dep}
	\depedge{3}{5}{ccomp}
	\depedge{10}{6}{mark}
	\depedge{8}{7}{det}
	\depedge{10}{8}{nsubj}
	\depedge{10}{9}{cop}
	\depedge{5}{10}{ccomp}
	\depedge[hide label, edge below, edge style={-, blue, opacity=0.5}]{4}{5}{}
	\depedge[hide label, edge below, edge style={-, blue, opacity=0.5}]{7}{8}{}
	\depedge[hide label, edge below, edge style={-, blue, opacity=0.5}]{2}{3}{}
	\depedge[hide label, edge below, edge style={-, blue, opacity=0.5}]{9}{10}{}
	\depedge[hide label, edge below, edge style={-, red, opacity=0.5}]{1}{2}{}
	\depedge[hide label, edge below, edge style={-, red, opacity=0.5}]{6}{8}{}
	\depedge[hide label, edge below, edge style={-, red, opacity=0.5}]{5}{6}{}
	\depedge[hide label, edge below, edge style={-, red, opacity=0.5}]{8}{9}{}
	\depedge[hide label, edge below, edge style={-, red, opacity=0.5}]{3}{4}{}
	\node (R) at (\matrixref.east) {{}};
	\node (R1) [right of = R] {\tiny\textsf{Bart}};
	\node (R4) at (R1.south) {\tiny $ 4/9 = 44\% $};
\end{dependency}

%% file: figures/tikz/ce/442sum.projective_BERT-base.tikz
\begin{dependency}
	\begin{deptext}
		It\& is\& impossible\& to\& know\& whether\& that\& theory\& is\& realistic\& . \\
	\end{deptext}
	\depedge[hide label, edge below, edge style={-, blue, opacity=0.5}]{7}{8}{}
	\depedge[hide label, edge below, edge style={-, blue, opacity=0.5}]{8}{10}{}
	\depedge[hide label, edge below, edge style={-, blue, opacity=0.5}]{2}{3}{}
	\depedge[hide label, edge below, edge style={-, red, opacity=0.5}]{6}{8}{}
	\depedge[hide label, edge below, edge style={-, red, opacity=0.5}]{5}{6}{}
	\depedge[hide label, edge below, edge style={-, red, opacity=0.5}]{8}{9}{}
	\depedge[edge unit distance=0.5ex,
	hide label, edge below, edge style={-, red, opacity=0.5}]{1}{8}{}
	\depedge[hide label, edge below, edge style={-, red, opacity=0.5}]{3}{6}{}
	\depedge[hide label, edge below, edge style={-, red, opacity=0.5}]{3}{4}{}
	\node (R) at (\matrixref.east) {{}};
	\node (R1) [right of = R] {\tiny\textsf{BERT-base}};
	\node (R4) at (R1.south) {\tiny $ 3/9 = 33\% $};
\end{dependency}

%% file: figures/tikz/ce/442sum.projective_DistilBERT.tikz
\begin{dependency}
	\begin{deptext}
		It\& is\& impossible\& to\& know\& whether\& that\& theory\& is\& realistic\& . \\
	\end{deptext}
	\depedge[hide label, edge below, edge style={-, blue, opacity=0.5}]{4}{5}{}
	\depedge[hide label, edge below, edge style={-, blue, opacity=0.5}]{6}{10}{}
	\depedge[hide label, edge below, edge style={-, blue, opacity=0.5}]{2}{3}{}
	\depedge[hide label, edge below, edge style={-, blue, opacity=0.5}]{9}{10}{}
	\depedge[hide label, edge below, edge style={-, red, opacity=0.5}]{1}{2}{}
	\depedge[hide label, edge below, edge style={-, red, opacity=0.5}]{6}{7}{}
	\depedge[hide label, edge below, edge style={-, red, opacity=0.5}]{6}{8}{}
	\depedge[hide label, edge below, edge style={-, red, opacity=0.5}]{5}{6}{}
	\depedge[hide label, edge below, edge style={-, red, opacity=0.5}]{3}{6}{}
	\node (R) at (\matrixref.east) {{}};
	\node (R1) [right of = R] {\tiny\textsf{DistilBERT}};
	\node (R4) at (R1.south) {\tiny $ 4/9 = 44\% $};
\end{dependency}

%% file: figures/tikz/ce/442sum.projective_XLM.tikz
\begin{dependency}
	\begin{deptext}
		It\& is\& impossible\& to\& know\& whether\& that\& theory\& is\& realistic\& . \\
	\end{deptext}
	\depedge[hide label, edge below, edge style={-, blue, opacity=0.5}]{6}{10}{}
	\depedge[hide label, edge below, edge style={-, blue, opacity=0.5}]{8}{10}{}
	\depedge[hide label, edge below, edge style={-, blue, opacity=0.5}]{2}{3}{}
	\depedge[hide label, edge below, edge style={-, blue, opacity=0.5}]{7}{8}{}
	\depedge[hide label, edge below, edge style={-, blue, opacity=0.5}]{3}{5}{}
	\depedge[hide label, edge below, edge style={-, red, opacity=0.5}]{5}{6}{}
	\depedge[hide label, edge below, edge style={-, red, opacity=0.5}]{1}{5}{}
	\depedge[hide label, edge below, edge style={-, red, opacity=0.5}]{3}{4}{}
	\depedge[hide label, edge below, edge style={-, red, opacity=0.5}]{8}{9}{}
	\node (R) at (\matrixref.east) {{}};
	\node (R1) [right of = R] {\tiny\textsf{XLM}};
	\node (R4) at (R1.south) {\tiny $ 5/9 = 56\% $};
\end{dependency}

%% file: figures/tikz/ce/442sum.projective_XLNet-base.tikz
\begin{dependency}
	\begin{deptext}
		It\& is\& impossible\& to\& know\& whether\& that\& theory\& is\& realistic\& . \\
	\end{deptext}
	\depedge[hide label, edge below, edge style={-, blue, opacity=0.5}]{4}{5}{}
	\depedge[hide label, edge below, edge style={-, blue, opacity=0.5}]{6}{10}{}
	\depedge[hide label, edge below, edge style={-, blue, opacity=0.5}]{8}{10}{}
	\depedge[hide label, edge below, edge style={-, blue, opacity=0.5}]{9}{10}{}
	\depedge[hide label, edge below, edge style={-, blue, opacity=0.5}]{2}{3}{}
	\depedge[hide label, edge below, edge style={-, blue, opacity=0.5}]{7}{8}{}
	\depedge[hide label, edge below, edge style={-, blue, opacity=0.5}]{3}{5}{}
	\depedge[hide label, edge below, edge style={-, red, opacity=0.5}]{1}{2}{}
	\depedge[hide label, edge below, edge style={-, red, opacity=0.5}]{5}{6}{}
	\node (R) at (\matrixref.east) {{}};
	\node (R1) [right of = R] {\tiny\textsf{XLNet-base}};
	\node (R4) at (R1.south) {\tiny $ 7/9 = 78\% $};
\end{dependency}

%% file: figures/tikz/ce/442sum.projective_W2V-signed.tikz
\begin{dependency}
	\begin{deptext}
		It\& is\& impossible\& to\& know\& whether\& that\& theory\& is\& realistic\& . \\
	\end{deptext}
	\depedge[hide label, edge below, edge style={-, blue, opacity=0.5}]{9}{10}{}
	\depedge[hide label, edge below, edge style={-, blue, opacity=0.5}]{7}{8}{}
	\depedge[hide label, edge below, edge style={-, red, opacity=0.5}]{1}{2}{}
	\depedge[hide label, edge below, edge style={-, red, opacity=0.5}]{6}{9}{}
	\depedge[hide label, edge below, edge style={-, red, opacity=0.5}]{5}{6}{}
	\depedge[hide label, edge below, edge style={-, red, opacity=0.5}]{8}{9}{}
	\depedge[edge unit distance=0.45ex,
	hide label, edge below, edge style={-, red, opacity=0.5}]{1}{9}{}
	\depedge[hide label, edge below, edge style={-, red, opacity=0.5}]{3}{6}{}
	\depedge[hide label, edge below, edge style={-, red, opacity=0.5}]{3}{4}{}
	\node (R) at (\matrixref.east) {{}};
	\node (R1) [right of = R] {\tiny\textsf{W2V-signed}};
	\node (R4) at (R1.south) {\tiny $ 2/9 = 22\% $};
\end{dependency}

%% file: figures/tikz/ce/1352sum.projective_Bart-wgold.tikz
\begin{dependency}
	\begin{deptext}
		Gold\& was\& nowhere\& the\& spectacular\& performer\& it\& was\& two\& years\& ago\& on\& Black\& Monday\& . \\
	\end{deptext}
	\depedge{6}{1}{nsubj}
	\depedge{6}{2}{cop}
	\depedge{6}{3}{advmod}
	\depedge{6}{4}{det}
	\depedge{6}{5}{amod}
	\depedge{8}{7}{nsubj}
	\depedge{6}{8}{rcmod}
	\depedge{10}{9}{num}
	\depedge{11}{10}{npadvmod}
	\depedge{8}{11}{advmod}
	\depedge{8}{12}{prep}
	\depedge{14}{13}{nn}
	\depedge{12}{14}{pobj}
	\depedge[hide label, edge below, edge style={-, blue, opacity=0.5}]{10}{11}{}
	\depedge[hide label, edge below, edge style={-, blue, opacity=0.5}]{4}{6}{}
	\depedge[hide label, edge below, edge style={-, blue, opacity=0.5}]{5}{6}{}
	\depedge[hide label, edge below, edge style={-, blue, opacity=0.5}]{13}{14}{}
	\depedge[hide label, edge below, edge style={-, blue, opacity=0.5}]{9}{10}{}
	\depedge[hide label, edge below, edge style={-, blue, opacity=0.5}]{7}{8}{}
	\depedge[hide label, edge below, edge style={-, red, opacity=0.5}]{1}{2}{}
	\depedge[hide label, edge below, edge style={-, red, opacity=0.5}]{4}{7}{}
	\depedge[edge unit distance=0.45ex,
	hide label, edge below, edge style={-, red, opacity=0.5}]{3}{13}{}
	\depedge[hide label, edge below, edge style={-, red, opacity=0.5}]{12}{13}{}
	\depedge[hide label, edge below, edge style={-, red, opacity=0.5}]{8}{9}{}
	\depedge[hide label, edge below, edge style={-, red, opacity=0.5}]{2}{3}{}
	\depedge[hide label, edge below, edge style={-, red, opacity=0.5}]{11}{12}{}
	\node (R) at (\matrixref.east) {{}};
	\node (R1) [right of = R] {\tiny\textsf{Bart}};
	\node (R4) at (R1.south) {\tiny $ 6/13 = 46\% $};
\end{dependency}

%% file: figures/tikz/ce/1352sum.projective_BERT-base.tikz
\begin{dependency}
	\begin{deptext}
		Gold\& was\& nowhere\& the\& spectacular\& performer\& it\& was\& two\& years\& ago\& on\& Black\& Monday\& . \\
	\end{deptext}
	\depedge[hide label, edge below, edge style={-, blue, opacity=0.5}]{10}{11}{}
	\depedge[hide label, edge below, edge style={-, blue, opacity=0.5}]{12}{14}{}
	\depedge[hide label, edge below, edge style={-, blue, opacity=0.5}]{4}{6}{}
	\depedge[hide label, edge below, edge style={-, blue, opacity=0.5}]{5}{6}{}
	\depedge[hide label, edge below, edge style={-, blue, opacity=0.5}]{13}{14}{}
	\depedge[hide label, edge below, edge style={-, blue, opacity=0.5}]{9}{10}{}
	\depedge[hide label, edge below, edge style={-, blue, opacity=0.5}]{1}{6}{}
	\depedge[hide label, edge below, edge style={-, blue, opacity=0.5}]{3}{6}{}
	\depedge[hide label, edge below, edge style={-, blue, opacity=0.5}]{7}{8}{}
	\depedge[hide label, edge below, edge style={-, red, opacity=0.5}]{6}{11}{}
	\depedge[hide label, edge below, edge style={-, red, opacity=0.5}]{11}{12}{}
	\depedge[hide label, edge below, edge style={-, red, opacity=0.5}]{6}{7}{}
	\depedge[hide label, edge below, edge style={-, red, opacity=0.5}]{2}{3}{}
	\node (R) at (\matrixref.east) {{}};
	\node (R1) [right of = R] {\tiny\textsf{BERT-base}};
	\node (R4) at (R1.south) {\tiny $ 9/13 = 69\% $};
\end{dependency}

%% file: figures/tikz/ce/1352sum.projective_XLM.tikz
\begin{dependency}
	\begin{deptext}
		Gold\& was\& nowhere\& the\& spectacular\& performer\& it\& was\& two\& years\& ago\& on\& Black\& Monday\& . \\
	\end{deptext}
	\depedge[hide label, edge below, edge style={-, blue, opacity=0.5}]{6}{8}{}
	\depedge[hide label, edge below, edge style={-, blue, opacity=0.5}]{4}{6}{}
	\depedge[hide label, edge below, edge style={-, blue, opacity=0.5}]{5}{6}{}
	\depedge[hide label, edge below, edge style={-, blue, opacity=0.5}]{13}{14}{}
	\depedge[hide label, edge below, edge style={-, blue, opacity=0.5}]{1}{6}{}
	\depedge[hide label, edge below, edge style={-, blue, opacity=0.5}]{3}{6}{}
	\depedge[hide label, edge below, edge style={-, red, opacity=0.5}]{9}{13}{}
	\depedge[hide label, edge below, edge style={-, red, opacity=0.5}]{6}{7}{}
	\depedge[hide label, edge below, edge style={-, red, opacity=0.5}]{12}{13}{}
	\depedge[hide label, edge below, edge style={-, red, opacity=0.5}]{10}{13}{}
	\depedge[hide label, edge below, edge style={-, red, opacity=0.5}]{2}{3}{}
	\depedge[hide label, edge below, edge style={-, red, opacity=0.5}]{11}{12}{}
	\depedge[hide label, edge below, edge style={-, red, opacity=0.5}]{6}{14}{}
	\node (R) at (\matrixref.east) {{}};
	\node (R1) [right of = R] {\tiny\textsf{XLM}};
	\node (R4) at (R1.south) {\tiny $ 6/13 = 46\% $};
\end{dependency}

%% file: figures/tikz/ce/1352sum.projective_W2V-signed.tikz
\begin{dependency}
	\begin{deptext}
		Gold\& was\& nowhere\& the\& spectacular\& performer\& it\& was\& two\& years\& ago\& on\& Black\& Monday\& . \\
	\end{deptext}
	\depedge[edge unit distance=0.55ex,
	hide label, edge below, edge style={-, blue, opacity=0.5}]{2}{6}{}
	\depedge[hide label, edge below, edge style={-, blue, opacity=0.5}]{10}{11}{}
	\depedge[hide label, edge below, edge style={-, blue, opacity=0.5}]{4}{6}{}
	\depedge[hide label, edge below, edge style={-, blue, opacity=0.5}]{5}{6}{}
	\depedge[hide label, edge below, edge style={-, blue, opacity=0.5}]{9}{10}{}
	\depedge[hide label, edge below, edge style={-, blue, opacity=0.5}]{7}{8}{}
	\depedge[edge unit distance=0.55ex,
	hide label, edge below, edge style={-, red, opacity=0.5}]{2}{7}{}
	\depedge[edge unit distance=0.35ex,
	hide label, edge below, edge style={-, red, opacity=0.5}]{1}{12}{}
	\depedge[edge unit distance=0.35ex,
	hide label, edge below, edge style={-, red, opacity=0.5}]{1}{13}{}
	\depedge[edge unit distance=0.35ex,
	hide label, edge below, edge style={-, red, opacity=0.5}]{1}{14}{}
	\depedge[edge unit distance=0.55ex,
	hide label, edge below, edge style={-, red, opacity=0.5}]{8}{10}{}
	\depedge[hide label, edge below, edge style={-, red, opacity=0.5}]{2}{3}{}
	\depedge[edge unit distance=0.35ex,
	hide label, edge below, edge style={-, red, opacity=0.5}]{2}{12}{}
	\node (R) at (\matrixref.east) {{}};
	\node (R1) [right of = R] {\tiny\textsf{W2V-signed}};
	\node (R4) at (R1.south) {\tiny $ 6/13 = 46\% $};
\end{dependency}

%% file: figures/tikz/ce/405sum.projective_Bart-wgold.tikz
\begin{dependency}
	\begin{deptext}
		``\& We\& have\& sufficient\& cash\& flow\& to\& handle\& that\& ,\& ''\& he\& said\& . \\
	\end{deptext}
	\depedge{3}{2}{nsubj}
	\depedge[edge unit distance=0.45ex]{13}{3}{ccomp}
	\depedge{6}{4}{amod}
	\depedge{6}{5}{nn}
	\depedge{3}{6}{dobj}
	\depedge{8}{7}{aux}
	\depedge{6}{8}{vmod}
	\depedge{8}{9}{dobj}
	\depedge{13}{12}{nsubj}
	\depedge[hide label, edge below, edge style={-, blue, opacity=0.5}]{5}{6}{}
	\depedge[hide label, edge below, edge style={-, blue, opacity=0.5}]{7}{8}{}
	\depedge[hide label, edge below, edge style={-, blue, opacity=0.5}]{12}{13}{}
	\depedge[hide label, edge below, edge style={-, blue, opacity=0.5}]{2}{3}{}
	\depedge[hide label, edge below, edge style={-, red, opacity=0.5}]{9}{13}{}
	\depedge[hide label, edge below, edge style={-, red, opacity=0.5}]{4}{9}{}
	\depedge[hide label, edge below, edge style={-, red, opacity=0.5}]{6}{7}{}
	\depedge[hide label, edge below, edge style={-, red, opacity=0.5}]{4}{5}{}
	\depedge[hide label, edge below, edge style={-, red, opacity=0.5}]{3}{4}{}
	\node (R) at (\matrixref.east) {{}};
	\node (R1) [right of = R] {\tiny\textsf{Bart}};
	\node (R4) at (R1.south) {\tiny $ 4/9 = 44\% $};
\end{dependency}

%% file: figures/tikz/ce/405sum.projective_BERT-base.tikz
\begin{dependency}
	\begin{deptext}
		``\& We\& have\& sufficient\& cash\& flow\& to\& handle\& that\& ,\& ''\& he\& said\& . \\
	\end{deptext}
	\depedge[hide label, edge below, edge style={-, blue, opacity=0.5}]{4}{6}{}
	\depedge[hide label, edge below, edge style={-, blue, opacity=0.5}]{5}{6}{}
	\depedge[hide label, edge below, edge style={-, blue, opacity=0.5}]{12}{13}{}
	\depedge[hide label, edge below, edge style={-, blue, opacity=0.5}]{2}{3}{}
	\depedge[hide label, edge below, edge style={-, blue, opacity=0.5}]{7}{8}{}
	\depedge[hide label, edge below, edge style={-, red, opacity=0.5}]{3}{4}{}
	\depedge[hide label, edge below, edge style={-, red, opacity=0.5}]{6}{7}{}
	\depedge[edge unit distance=0.45ex,
	hide label, edge below, edge style={-, red, opacity=0.5}]{2}{12}{}
	\depedge[edge unit distance=0.45ex,
	hide label, edge below, edge style={-, red, opacity=0.5}]{2}{9}{}
	\node (R) at (\matrixref.east) {{}};
	\node (R1) [right of = R] {\tiny\textsf{BERT-base}};
	\node (R4) at (R1.south) {\tiny $ 5/9 = 56\% $};
\end{dependency}

%% file: figures/tikz/ce/405sum.projective_DistilBERT.tikz
\begin{dependency}
	\begin{deptext}
		``\& We\& have\& sufficient\& cash\& flow\& to\& handle\& that\& ,\& ''\& he\& said\& . \\
	\end{deptext}
	\depedge[hide label, edge below, edge style={-, blue, opacity=0.5}]{4}{6}{}
	\depedge[hide label, edge below, edge style={-, blue, opacity=0.5}]{5}{6}{}
	\depedge[hide label, edge below, edge style={-, blue, opacity=0.5}]{12}{13}{}
	\depedge[hide label, edge below, edge style={-, blue, opacity=0.5}]{8}{9}{}
	\depedge[hide label, edge below, edge style={-, blue, opacity=0.5}]{2}{3}{}
	\depedge[hide label, edge below, edge style={-, blue, opacity=0.5}]{7}{8}{}
	\depedge[hide label, edge below, edge style={-, red, opacity=0.5}]{3}{4}{}
	\depedge[hide label, edge below, edge style={-, red, opacity=0.5}]{4}{7}{}
	\depedge[hide label, edge below, edge style={-, red, opacity=0.5}]{8}{13}{}
	\node (R) at (\matrixref.east) {{}};
	\node (R1) [right of = R] {\tiny\textsf{DistilBERT}};
	\node (R4) at (R1.south) {\tiny $ 6/9 = 67\% $};
\end{dependency}

%% file: figures/tikz/ce/405sum.projective_XLM.tikz
\begin{dependency}
	\begin{deptext}
		``\& We\& have\& sufficient\& cash\& flow\& to\& handle\& that\& ,\& ''\& he\& said\& . \\
	\end{deptext}
	\depedge[hide label, edge below, edge style={-, blue, opacity=0.5}]{5}{6}{}
	\depedge[hide label, edge below, edge style={-, blue, opacity=0.5}]{7}{8}{}
	\depedge[hide label, edge below, edge style={-, blue, opacity=0.5}]{4}{6}{}
	\depedge[hide label, edge below, edge style={-, blue, opacity=0.5}]{2}{3}{}
	\depedge[edge unit distance=0.6ex,
	hide label, edge below, edge style={-, red, opacity=0.5}]{4}{8}{}
	\depedge[edge unit distance=0.45ex,
	hide label, edge below, edge style={-, red, opacity=0.5}]{4}{13}{}
	\depedge[edge unit distance=0.45ex,
	hide label, edge below, edge style={-, red, opacity=0.5}]{4}{12}{}
	\depedge[hide label, edge below, edge style={-, red, opacity=0.5}]{3}{4}{}
	\depedge[edge unit distance=0.45ex,
	hide label, edge below, edge style={-, red, opacity=0.5}]{9}{12}{}
	\node (R) at (\matrixref.east) {{}};
	\node (R1) [right of = R] {\tiny\textsf{XLM}};
	\node (R4) at (R1.south) {\tiny $ 4/9 = 44\% $};
\end{dependency}

%% file: figures/tikz/ce/405sum.projective_XLNet-base.tikz
\begin{dependency}
	\begin{deptext}
		``\& We\& have\& sufficient\& cash\& flow\& to\& handle\& that\& ,\& ''\& he\& said\& . \\
	\end{deptext}
	\depedge[hide label, edge below, edge style={-, blue, opacity=0.5}]{4}{6}{}
	\depedge[hide label, edge below, edge style={-, blue, opacity=0.5}]{5}{6}{}
	\depedge[hide label, edge below, edge style={-, blue, opacity=0.5}]{12}{13}{}
	\depedge[hide label, edge below, edge style={-, blue, opacity=0.5}]{2}{3}{}
	\depedge[hide label, edge below, edge style={-, blue, opacity=0.5}]{7}{8}{}
	\depedge[hide label, edge below, edge style={-, red, opacity=0.5}]{2}{6}{}
	\depedge[hide label, edge below, edge style={-, red, opacity=0.5}]{6}{7}{}
	\depedge[hide label, edge below, edge style={-, red, opacity=0.5}]{9}{12}{}
	\depedge[hide label, edge below, edge style={-, red, opacity=0.5}]{7}{9}{}
	\node (R) at (\matrixref.east) {{}};
	\node (R1) [right of = R] {\tiny\textsf{XLNet-base}};
	\node (R4) at (R1.south) {\tiny $ 5/9 = 56\% $};
\end{dependency}

%% file: figures/tikz/ce/405sum.projective_W2V-signed.tikz
\begin{dependency}
	\begin{deptext}
		``\& We\& have\& sufficient\& cash\& flow\& to\& handle\& that\& ,\& ''\& he\& said\& . \\
	\end{deptext}
	\depedge[hide label, edge below, edge style={-, blue, opacity=0.5}]{4}{6}{}
	\depedge[edge unit distance=0.4ex,
	hide label, edge below, edge style={-, blue, opacity=0.5}]{3}{13}{}
	\depedge[hide label, edge below, edge style={-, blue, opacity=0.5}]{5}{6}{}
	\depedge[hide label, edge below, edge style={-, blue, opacity=0.5}]{12}{13}{}
	\depedge[hide label, edge below, edge style={-, blue, opacity=0.5}]{2}{3}{}
	\depedge[edge unit distance=0.45ex,
	hide label, edge below, edge style={-, red, opacity=0.5}]{9}{13}{}
	\depedge[hide label, edge below, edge style={-, red, opacity=0.5}]{4}{7}{}
	\depedge[hide label, edge below, edge style={-, red, opacity=0.5}]{3}{4}{}
	\depedge[hide label, edge below, edge style={-, red, opacity=0.5}]{4}{8}{}
	\node (R) at (\matrixref.east) {{}};
	\node (R1) [right of = R] {\tiny\textsf{W2V-signed}};
	\node (R4) at (R1.south) {\tiny $ 5/9 = 56\% $};
\end{dependency}

%% file: figures/tikz/lstm/442sum.projective_LSTM-wgold.tikz
\begin{dependency}
	\begin{deptext}
		It\& is\& impossible\& to\& know\& whether\& that\& theory\& is\& realistic\& . \\
	\end{deptext}
	\depedge{3}{1}{nsubj}
	\depedge{3}{2}{cop}
	\depedge{5}{4}{dep}
	\depedge{3}{5}{ccomp}
	\depedge{10}{6}{mark}
	\depedge{8}{7}{det}
	\depedge{10}{8}{nsubj}
	\depedge{10}{9}{cop}
	\depedge{5}{10}{ccomp}
	\depedge[hide label, edge below, edge style={-, blue, opacity=0.5}]{1}{3}{}
	\depedge[hide label, edge below, edge style={-, blue, opacity=0.5}]{6}{10}{}
	\depedge[hide label, edge below, edge style={-, blue, opacity=0.5}]{9}{10}{}
	\depedge[hide label, edge below, edge style={-, blue, opacity=0.5}]{7}{8}{}
	\depedge[hide label, edge below, edge style={-, blue, opacity=0.5}]{3}{5}{}
	\depedge[hide label, edge below, edge style={-, red, opacity=0.5}]{1}{2}{}
	\depedge[hide label, edge below, edge style={-, red, opacity=0.5}]{8}{9}{}
	\depedge[hide label, edge below, edge style={-, red, opacity=0.5}]{3}{4}{}
	\depedge[hide label, edge below, edge style={-, red, opacity=0.5}]{5}{6}{}
	\node (R) at (\matrixref.east) {{}};
	\node (R1) [right of = R] {\tiny\textsf{LSTM}};
	\node (R4) at (R1.south) {\tiny $ 5/9 = 56\% $};
\end{dependency}

%% file: figures/tikz/lstm/442sum.projective_ONLSTM.tikz
\begin{dependency}
	\begin{deptext}
		It\& is\& impossible\& to\& know\& whether\& that\& theory\& is\& realistic\& . \\
	\end{deptext}
	\depedge[hide label, edge below, edge style={-, blue, opacity=0.5}]{9}{10}{}
	\depedge[hide label, edge below, edge style={-, blue, opacity=0.5}]{1}{3}{}
	\depedge[hide label, edge below, edge style={-, blue, opacity=0.5}]{3}{5}{}
	\depedge[hide label, edge below, edge style={-, red, opacity=0.5}]{1}{2}{}
	\depedge[hide label, edge below, edge style={-, red, opacity=0.5}]{6}{7}{}
	\depedge[hide label, edge below, edge style={-, red, opacity=0.5}]{6}{8}{}
	\depedge[hide label, edge below, edge style={-, red, opacity=0.5}]{5}{6}{}
	\depedge[hide label, edge below, edge style={-, red, opacity=0.5}]{8}{9}{}
	\depedge[hide label, edge below, edge style={-, red, opacity=0.5}]{3}{4}{}
	\node (R) at (\matrixref.east) {{}};
	\node (R1) [right of = R] {\tiny\textsf{ONLSTM}};
	\node (R4) at (R1.south) {\tiny $ 3/9 = 33\% $};
\end{dependency}

%% file: figures/tikz/lstm/442sum.projective_ONLSTM-SYD.tikz
\begin{dependency}
	\begin{deptext}
		It\& is\& impossible\& to\& know\& whether\& that\& theory\& is\& realistic\& . \\
	\end{deptext}
	\depedge[hide label, edge below, edge style={-, blue, opacity=0.5}]{9}{10}{}
	\depedge[hide label, edge below, edge style={-, blue, opacity=0.5}]{1}{3}{}
	\depedge[hide label, edge below, edge style={-, blue, opacity=0.5}]{7}{8}{}
	\depedge[hide label, edge below, edge style={-, blue, opacity=0.5}]{3}{5}{}
	\depedge[hide label, edge below, edge style={-, red, opacity=0.5}]{1}{2}{}
	\depedge[hide label, edge below, edge style={-, red, opacity=0.5}]{6}{9}{}
	\depedge[hide label, edge below, edge style={-, red, opacity=0.5}]{5}{6}{}
	\depedge[hide label, edge below, edge style={-, red, opacity=0.5}]{8}{9}{}
	\depedge[hide label, edge below, edge style={-, red, opacity=0.5}]{3}{4}{}
	\node (R) at (\matrixref.east) {{}};
	\node (R1) [right of = R] {\tiny\textsf{ONLSTM-SYD}};
	\node (R4) at (R1.south) {\tiny $ 4/9 = 44\% $};
\end{dependency}

%% file: figures/tikz/lstm/1352sum.projective_LSTM-wgold.tikz
\begin{dependency}
	\begin{deptext}
		Gold\& was\& nowhere\& the\& spectacular\& performer\& it\& was\& two\& years\& ago\& on\& Black\& Monday\& . \\
	\end{deptext}
	\depedge{6}{1}{nsubj}
	\depedge{6}{2}{cop}
	\depedge{6}{3}{advmod}
	\depedge{6}{4}{det}
	\depedge{6}{5}{amod}
	\depedge{8}{7}{nsubj}
	\depedge{6}{8}{rcmod}
	\depedge{10}{9}{num}
	\depedge{11}{10}{npadvmod}
	\depedge{8}{11}{advmod}
	\depedge{8}{12}{prep}
	\depedge{14}{13}{nn}
	\depedge{12}{14}{pobj}
	\depedge[hide label, edge below, edge style={-, blue, opacity=0.5}]{12}{14}{}
	\depedge[hide label, edge below, edge style={-, blue, opacity=0.5}]{4}{6}{}
	\depedge[hide label, edge below, edge style={-, blue, opacity=0.5}]{5}{6}{}
	\depedge[hide label, edge below, edge style={-, blue, opacity=0.5}]{13}{14}{}
	\depedge[hide label, edge below, edge style={-, blue, opacity=0.5}]{9}{10}{}
	\depedge[hide label, edge below, edge style={-, blue, opacity=0.5}]{7}{8}{}
	\depedge[hide label, edge below, edge style={-, red, opacity=0.5}]{1}{2}{}
	\depedge[hide label, edge below, edge style={-, red, opacity=0.5}]{2}{8}{}
	\depedge[hide label, edge below, edge style={-, red, opacity=0.5}]{8}{10}{}
	\depedge[hide label, edge below, edge style={-, red, opacity=0.5}]{10}{12}{}
	\depedge[hide label, edge below, edge style={-, red, opacity=0.5}]{2}{3}{}
	\depedge[hide label, edge below, edge style={-, red, opacity=0.5}]{3}{4}{}
	\depedge[hide label, edge below, edge style={-, red, opacity=0.5}]{11}{12}{}
	\node (R) at (\matrixref.east) {{}};
	\node (R1) [right of = R] {\tiny\textsf{LSTM}};
	\node (R4) at (R1.south) {\tiny $ 6/13 = 46\% $};
\end{dependency}

%% file: figures/tikz/lstm/1352sum.projective_ONLSTM.tikz
\begin{dependency}
	\begin{deptext}
		Gold\& was\& nowhere\& the\& spectacular\& performer\& it\& was\& two\& years\& ago\& on\& Black\& Monday\& . \\
	\end{deptext}
	\depedge[hide label, edge below, edge style={-, blue, opacity=0.5}]{12}{14}{}
	\depedge[hide label, edge below, edge style={-, blue, opacity=0.5}]{4}{6}{}
	\depedge[hide label, edge below, edge style={-, blue, opacity=0.5}]{5}{6}{}
	\depedge[hide label, edge below, edge style={-, blue, opacity=0.5}]{13}{14}{}
	\depedge[hide label, edge below, edge style={-, blue, opacity=0.5}]{9}{10}{}
	\depedge[hide label, edge below, edge style={-, blue, opacity=0.5}]{7}{8}{}
	\depedge[hide label, edge below, edge style={-, red, opacity=0.5}]{1}{2}{}
	\depedge[edge unit distance=0.45ex,
	hide label, edge below, edge style={-, red, opacity=0.5}]{2}{8}{}
	\depedge[edge unit distance=0.45ex,
	hide label, edge below, edge style={-, red, opacity=0.5}]{1}{14}{}
	\depedge[hide label, edge below, edge style={-, red, opacity=0.5}]{10}{14}{}
	\depedge[hide label, edge below, edge style={-, red, opacity=0.5}]{2}{3}{}
	\depedge[hide label, edge below, edge style={-, red, opacity=0.5}]{11}{14}{}
	\depedge[hide label, edge below, edge style={-, red, opacity=0.5}]{3}{4}{}
	\node (R) at (\matrixref.east) {{}};
	\node (R1) [right of = R] {\tiny\textsf{ONLSTM}};
	\node (R4) at (R1.south) {\tiny $ 6/13 = 46\% $};
\end{dependency}

%% file: figures/tikz/lstm/1352sum.projective_ONLSTM-SYD.tikz
\begin{dependency}
	\begin{deptext}
		Gold\& was\& nowhere\& the\& spectacular\& performer\& it\& was\& two\& years\& ago\& on\& Black\& Monday\& . \\
	\end{deptext}
	\depedge[hide label, edge below, edge style={-, blue, opacity=0.5}]{10}{11}{}
	\depedge[hide label, edge below, edge style={-, blue, opacity=0.5}]{12}{14}{}
	\depedge[hide label, edge below, edge style={-, blue, opacity=0.5}]{4}{6}{}
	\depedge[hide label, edge below, edge style={-, blue, opacity=0.5}]{5}{6}{}
	\depedge[hide label, edge below, edge style={-, blue, opacity=0.5}]{13}{14}{}
	\depedge[hide label, edge below, edge style={-, blue, opacity=0.5}]{9}{10}{}
	\depedge[hide label, edge below, edge style={-, blue, opacity=0.5}]{7}{8}{}
	\depedge[hide label, edge below, edge style={-, red, opacity=0.5}]{4}{7}{}
	\depedge[hide label, edge below, edge style={-, red, opacity=0.5}]{1}{4}{}
	\depedge[hide label, edge below, edge style={-, red, opacity=0.5}]{2}{3}{}
	\depedge[hide label, edge below, edge style={-, red, opacity=0.5}]{7}{10}{}
	\depedge[hide label, edge below, edge style={-, red, opacity=0.5}]{11}{14}{}
	\depedge[hide label, edge below, edge style={-, red, opacity=0.5}]{3}{4}{}
	\node (R) at (\matrixref.east) {{}};
	\node (R1) [right of = R] {\tiny\textsf{ONLSTM-SYD}};
	\node (R4) at (R1.south) {\tiny $ 7/13 = 54\% $};
\end{dependency}

%% file: figures/tikz/lstm/405sum.projective_LSTM-wgold.tikz
\begin{dependency}
	\begin{deptext}
		``\& We\& have\& sufficient\& cash\& flow\& to\& handle\& that\& ,\& ''\& he\& said\& . \\
	\end{deptext}
	\depedge{3}{2}{nsubj}
	\depedge[edge unit distance=0.45ex]{13}{3}{ccomp}
	\depedge{6}{4}{amod}
	\depedge{6}{5}{nn}
	\depedge{3}{6}{dobj}
	\depedge{8}{7}{aux}
	\depedge{6}{8}{vmod}
	\depedge{8}{9}{dobj}
	\depedge{13}{12}{nsubj}
	\depedge[hide label, edge below, edge style={-, blue, opacity=0.5}]{4}{6}{}
	\depedge[hide label, edge below, edge style={-, blue, opacity=0.5}]{5}{6}{}
	\depedge[hide label, edge below, edge style={-, blue, opacity=0.5}]{12}{13}{}
	\depedge[hide label, edge below, edge style={-, blue, opacity=0.5}]{2}{3}{}
	\depedge[hide label, edge below, edge style={-, blue, opacity=0.5}]{7}{8}{}
	\depedge[hide label, edge below, edge style={-, red, opacity=0.5}]{4}{7}{}
	\depedge[edge unit distance=0.45ex,
	hide label, edge below, edge style={-, red, opacity=0.5}]{2}{12}{}
	\depedge[hide label, edge below, edge style={-, red, opacity=0.5}]{9}{12}{}
	\depedge[hide label, edge below, edge style={-, red, opacity=0.5}]{2}{4}{}
	\node (R) at (\matrixref.east) {{}};
	\node (R1) [right of = R] {\tiny\textsf{LSTM}};
	\node (R4) at (R1.south) {\tiny $ 5/9 = 56\% $};
\end{dependency}

%% file: figures/tikz/lstm/405sum.projective_ONLSTM.tikz
\begin{dependency}
	\begin{deptext}
		``\& We\& have\& sufficient\& cash\& flow\& to\& handle\& that\& ,\& ''\& he\& said\& . \\
	\end{deptext}
	\depedge[hide label, edge below, edge style={-, blue, opacity=0.5}]{5}{6}{}
	\depedge[hide label, edge below, edge style={-, blue, opacity=0.5}]{2}{3}{}
	\depedge[hide label, edge below, edge style={-, blue, opacity=0.5}]{7}{8}{}
	\depedge[hide label, edge below, edge style={-, blue, opacity=0.5}]{12}{13}{}
	\depedge[hide label, edge below, edge style={-, red, opacity=0.5}]{6}{7}{}
	\depedge[hide label, edge below, edge style={-, red, opacity=0.5}]{4}{5}{}
	\depedge[edge unit distance=0.45ex,
	hide label, edge below, edge style={-, red, opacity=0.5}]{2}{12}{}
	\depedge[hide label, edge below, edge style={-, red, opacity=0.5}]{2}{4}{}
	\depedge[edge unit distance=0.45ex,
	hide label, edge below, edge style={-, red, opacity=0.5}]{9}{12}{}
	\node (R) at (\matrixref.east) {{}};
	\node (R1) [right of = R] {\tiny\textsf{ONLSTM}};
	\node (R4) at (R1.south) {\tiny $ 4/9 = 44\% $};
\end{dependency}

%% file: figures/tikz/lstm/405sum.projective_ONLSTM-SYD.tikz
\begin{dependency}
	\begin{deptext}
		``\& We\& have\& sufficient\& cash\& flow\& to\& handle\& that\& ,\& ''\& he\& said\& . \\
	\end{deptext}
	\depedge[hide label, edge below, edge style={-, blue, opacity=0.5}]{5}{6}{}
	\depedge[hide label, edge below, edge style={-, blue, opacity=0.5}]{2}{3}{}
	\depedge[hide label, edge below, edge style={-, blue, opacity=0.5}]{7}{8}{}
	\depedge[hide label, edge below, edge style={-, blue, opacity=0.5}]{12}{13}{}
	\depedge[hide label, edge below, edge style={-, red, opacity=0.5}]{6}{7}{}
	\depedge[hide label, edge below, edge style={-, red, opacity=0.5}]{8}{13}{}
	\depedge[hide label, edge below, edge style={-, red, opacity=0.5}]{4}{5}{}
	\depedge[hide label, edge below, edge style={-, red, opacity=0.5}]{2}{4}{}
	\depedge[hide label, edge below, edge style={-, red, opacity=0.5}]{9}{12}{}
	\node (R) at (\matrixref.east) {{}};
	\node (R1) [right of = R] {\tiny\textsf{ONLSTM-SYD}};
	\node (R4) at (R1.south) {\tiny $ 4/9 = 44\% $};
\end{dependency}